%% file: main.tex
\documentclass[11pt]{article}
	
	\newcommand{\blind}{0}
	
    \makeatletter
    \renewcommand\section{\@startsection {section}{1}{\z@}%
                                       {-3.5ex \@plus -1ex \@minus -.2ex}%
                                       {2.3ex \@plus.2ex}%
                                       {\normalfont\fontfamily{phv}\fontsize{16}{19}\bfseries}}
    \renewcommand\subsection{\@startsection{subsection}{2}{\z@}%
                                         {-3.25ex\@plus -1ex \@minus -.2ex}%
                                         {1.5ex \@plus .2ex}%
                                         {\normalfont\fontfamily{phv}\fontsize{14}{17}\bfseries}}
    \renewcommand\subsubsection{\@startsection{subsubsection}{3}{\z@}%
                                        {-3.25ex\@plus -1ex \@minus -.2ex}%
                                         {1.5ex \@plus .2ex}%
                                         {\normalfont\normalsize\fontfamily{phv}\fontsize{14}{17}\selectfont}}
    \makeatother
	
	\usepackage{amsmath}
	\usepackage{graphicx}
	\usepackage{enumerate}
	\usepackage{xcolor}
	\usepackage{natbib} 
	\usepackage{url} 
	
    \usepackage{amsmath}
    \usepackage{amssymb}
    \usepackage{amsthm}
    \newtheorem{definition}{Definition}[section] 
    \newtheorem{theorem}{Theorem}[section]
    \newtheorem{proposition}{Proposition}[section]
    \numberwithin{table}{section}
    
    \usepackage{algorithm}
    \usepackage{algorithmicx}
    \usepackage{algpseudocode}
    \floatname{algorithm}{Algorithm}

    \usepackage{enumitem}
    \usepackage{longtable}
    \usepackage{makecell}
    \usepackage{booktabs}
    \usepackage{tabularx}
    \usepackage{graphicx}
    \usepackage{float}
    \usepackage[hypcap=false]{caption}
    \usepackage[colorlinks=true, allcolors=blue]{hyperref}
    \usepackage{color}
    \usepackage{xcolor}
    \usepackage{soul}
    \usepackage[most]{tcolorbox} 
    \numberwithin{figure}{section}
    \numberwithin{algorithm}{section}
    \usepackage{appendix}
    \usepackage{cleveref}
    
    \usepackage{multirow}
    \usepackage{multicol}
    \usepackage{booktabs}
    \usepackage{makecell}
    
    \usepackage{algorithm}
    \usepackage{algorithmicx}
    \usepackage{algpseudocode}
    \floatname{algorithm}{Algorithm}

    \usepackage{soul, color, xcolor}

    \DeclareUnicodeCharacter{2061}{}
    \counterwithin{figure}{section}
    \counterwithin{table}{section}

    \makeatletter
    \newcommand{\rmnum}[1]{\romannumeral #1}
    \newcommand{\Rmnum}[1]{\expandafter\@slowromancap\romannumeral #1@}
    \makeatother

    \usepackage{natbib}
    \setcitestyle{authoryear,round}

    \usepackage{appendix}
    \usepackage{indentfirst}
    \allowdisplaybreaks[4]
	
\begin{document}
		
		\def\spacingset#1{\renewcommand{\baselinestretch}%
			{#1}\small\normalsize} \spacingset{1}
		
		\if0\blind
		{
			\title{\bf High Dimensional Data Decomposition for \\ Anomaly Detection of Textured Images}
			\author{Ji Song,\textsuperscript{a} 
            Xing Wang,\textsuperscript{b} Jianguo Wu,\textsuperscript{c} Xiaowei Yue,\textsuperscript{a}
            \\
            \thanks{\textsuperscript{a}Tsinghua University}
            \thanks{\textsuperscript{b}Illinois State University }
            \thanks{\textsuperscript{c}Peking University}
            }
            \author{Ji Song\textsuperscript{a}, 
            Xing Wang\textsuperscript{b}, Jianguo Wu\textsuperscript{c}, Xiaowei Yue\textsuperscript{a}*
            \\
            \textsuperscript{a}Tsinghua University \\
            \textsuperscript{b}Illinois State University \\
            \textsuperscript{c}Peking University\\
            \\
            (Corresponding Author: Xiaowei Yue, yuex@tsinghua.edu.cn)
            \footnote{This paper has been accepted by IISE Transactions. 
            \\ https://doi.org/10.1080/24725854.2025.2606372}
            }
			\date{}
			\maketitle
		} \fi
		
		\if1\blind
		{

            \title{\bf High Dimensional Data Decomposition for \\ Anomaly Detection of Textured Images}
			\author{Author information is purposely removed for double-blind review}
			
\bigskip
                \date{}
			\maketitle
		} \fi
		\bigskip
		
	\begin{abstract}
            In the realm of diverse high-dimensional data, images play a significant role across various processes of manufacturing systems where efficient image anomaly detection has emerged as a core technology of utmost importance.  
            However, when applied to textured defect images, conventional anomaly detection methods have limitations including non-negligible misidentification, low robustness, and excessive reliance on large-scale and structured datasets. This paper proposes a texture basis integrated smooth decomposition (TBSD) approach, which is targeted at efficient anomaly detection in textured images with smooth backgrounds and sparse anomalies. Mathematical formulation  of quasi-periodicity and its theoretical properties are investigated for image texture estimation. TBSD method consists of two principal processes: the first process learns the texture basis functions to effectively extract quasi-periodic texture patterns; the subsequent anomaly detection process utilizes that texture basis as prior knowledge to prevent texture misidentification and capture potential anomalies with high accuracy.
            The proposed method surpasses benchmarks with less misidentification, smaller training dataset requirement, and superior anomaly detection performance on both simulation and real-world datasets.
	\end{abstract}
			
	\noindent%
	{\it Keywords:} Quasi-periodicity, data decomposition, anomaly detection, image analysis.

	\spacingset{1.5} 


\section{Introduction} \label{s:intro}
\begin{figure}[h]
    \centering
    \includegraphics[width=\textwidth]{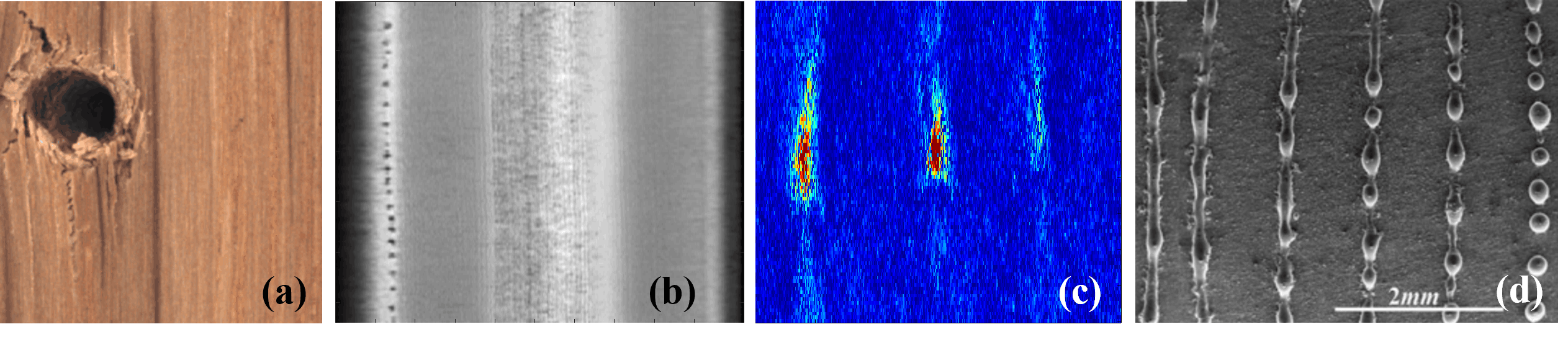}
    \caption{Real-world textured images in manufacturing: 
(a) Wood plate surface\citep{bergmann2021mvtec}, 
(b) Rolling steel inspection (from OG Technology), 
(c) Surface non-destructive testing (from Boeing), 
(d) 3D printing
}
    \label{fig:intro_ADapp}
\end{figure}

Image data, due to its universality, consistency, and rich feature information, has been widely used for storing and disseminating information. Image anomaly detection plays a critical role in modern manufacturing and service industries, especially with the growing application of lean production and smart factory by enterprises. As depicted in Figure \ref{fig:intro_ADapp}, the applicability of image anomaly detection is vast, involving wood production, steel rolling of high-speed railways, aviation material production, and 3D printing.
However, on account of accuracy and efficiency, which are highly valued by methods for image anomaly detection,
existing methods, such as Robust Principal Component Analysis (RPCA)\citep{candes2011robust}, Smooth-sparse Decomposition (SSD)\citep{yan2017anomaly}, and Fast Fourier Transform (FFT)\citep{tsai2003automated}, are not
fully capable of detecting anomalies with acceptable accuracy in scenarios with disturbance \textcolor{black}{(overlight, shadow, and camera shake, for example)}, unqualified
training dataset \textcolor{black}{(lacking image anomaly labels or valid defect images)} and especially mixed texture patterns \textcolor{black}{(indistinguishable from background or anomalies to a certain extent)}.

\begin{figure}[H]
    \centering
    \includegraphics[width=.7\linewidth]{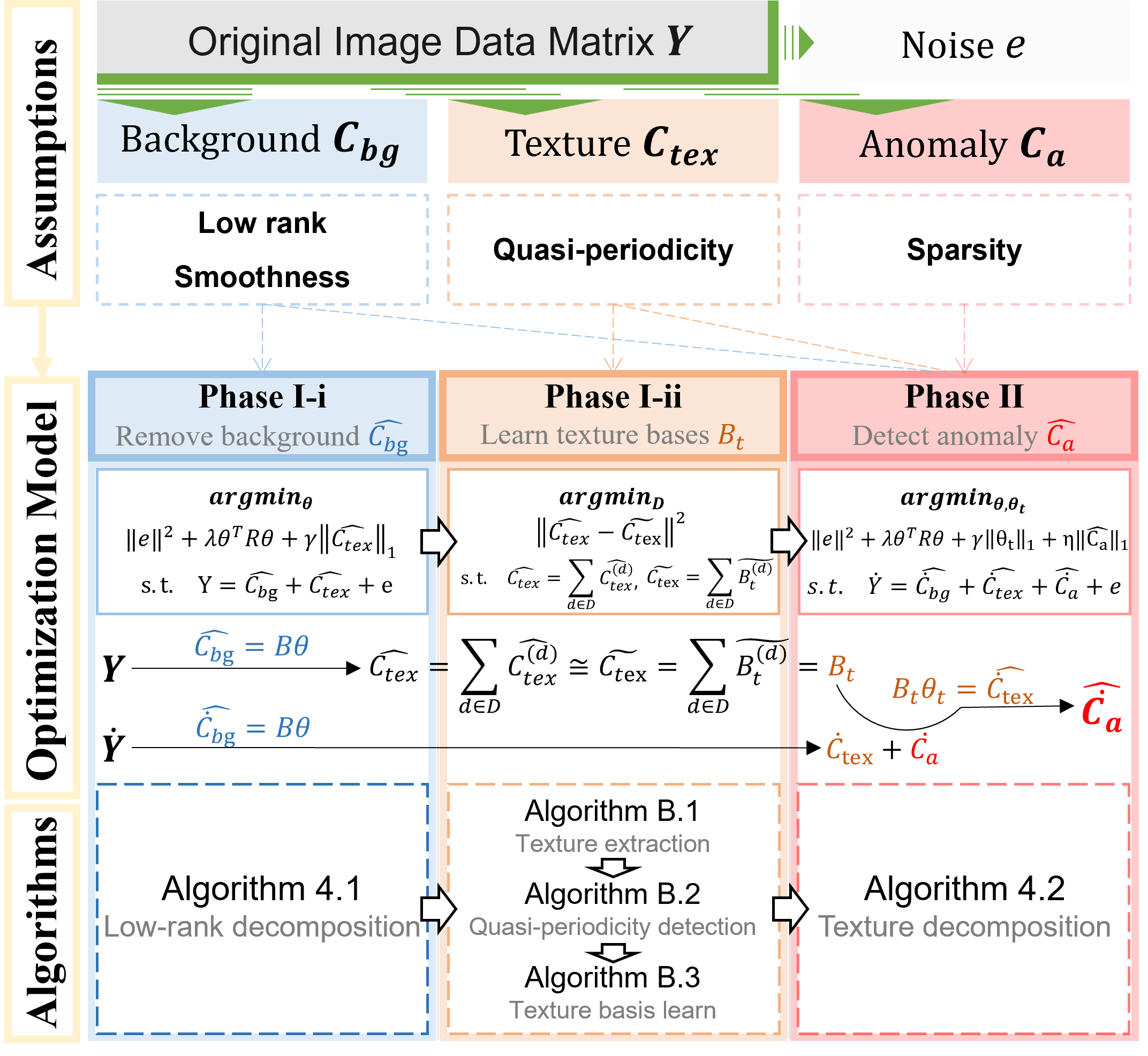}
    \caption{\textcolor{black}{Schematic diagram of relationships among Core mathematical assumptions, optimization models, and technical algorithms of the proposed TBSD method. 
    }}
    \label{fig:paper_structure}
\end{figure}

This paper proposes an efficient data decomposition driven anomaly detection method, named as Texture Basis Integrated Smooth Decomposition (TBSD), targetedly designed for textured images which consists of smooth background, quasi-periodic texture, and sparse anomalies.
Those image textures are assumed to possess quasi-periodicity, a characteristic which assumes there are common patterns within global textures. It is derived from periodicity and would be comprehensively discussed in Section \ref{s:perio}. 
The proposed TBSD method learns these common texture patterns as texture basis so as to effectively distinguish anomalies from textures without much reliance on high-quality training dataset. 
\textcolor{black}{
As shown in Figure \ref{fig:paper_structure}, the core optimization model and practical algorithms in Section \ref{s:method} are supported by the rigorous mathematical framework in Section \ref{s:perio}. Both simulation (Section \ref{s:simuexp}) and real-world experiments (Section \ref{s:realexp}) have sufficiently validated the efficiency of proposed approach.
}

The main novelty and contributions of this paper can be summarized as follows: 
\begin{enumerate}
      \item \textbf{This paper presents a theoretical analysis of quasi-periodicity in textured images.}
      The concept of quasi-periodicity is introduced to model how textures distribute in images to serve as the foundation of texture basis function learning.
      \item \textbf{This paper proposes an effective approach to prevent misidentifying normal textures as anomalies.}
      The core idea is to learn a series of texture basis functions that capture common texture patterns based on a small amount of defect-free images. The learned texture basis is subsequently utilized for image anomaly detection to prevent misidentification with stably high detection accuracy.
      \item \textbf{This paper has improved data decomposition by introducing prior knowledge to achieve better estimation of targeted components.}
      Conventional data decomposition can result in mixed output with simultaneous existence of similar data components.
      To further distinguish image textures and anomalies, prior knowledge of textures is introduced into the data decomposition model to obtain a better texture estimator, which helps restrain the model's overestimation of unknown anomalies. 
\end{enumerate}

\section{Literature Review} \label{s:liter}
\subsection{Related Methods for Textured Image Anomaly Detection}
\label{subs:liter_imgAD}
General anomaly detection methods could meet challenges for textured images due to their high dimensionality, large scale, and dataset imbalance. 
Diverse anomaly detection performance can be achieved according to model assumptions and algorithms chosen by alternative image detection methods, which include \textbf{statistical model based, domain transformation based, clustering based, image reconstruction based, and machine learning based methods}, as depicted in Table \ref{table: liter_imageADmethods}. 

\begin{table}[h]
    \centering
    \setlength{\tabcolsep}{3pt}
    \caption{Classification Example of image anomaly detection methods}
    \label{table: liter_imageADmethods}
    \begin{tabular}{|c|c|c|}
      \hline
      \textbf{Driven By} & \textbf{General Categories} & \textbf{Example} \\
      \hline
      \multirow{2}{*}{\textbf{\thead{Statistical \\ Model}}} & Parametric model based & Hierarchical Model\citep{Gao02092023} \\
      \cline{2-3}
       & Non-parametric model based & Parzen Windowing\citep{Veracini2011Nonparametric} \\
      \hline
      \multirow{2}{*}{\textbf{\thead{Domain \\ Transformation}}} & Spatial $\rightarrow$ Frequency & Fourier Transform\citep{tsai2003automated} \\
      \cline{2-3}
       & Spatial $\rightarrow$ other Domains & Hough Transform\citep{hma2006HoughTrans} \\
      \hline
      \multirow{2}{*}{\textbf{Clustering}} & Distance based & K-nearest Neighbors\citep{Tu2020HADusingDWD} \\
      \cline{2-3}
       & Density based & DBSCAN\citep{Hu2020localKDE} \\
      \hline
      \multirow{2}{*}{\textbf{\thead{Image \\ Reconstruction}}} & Low-rank Assumption based & Principal Component Analysis\citep{candes2011robust} \\
      \cline{2-3}
       & Sparsity Assumption based & Sparse Dictionary Learning\citep{li_anomalydetectionbased_2018} \\
      \hline
      \multirow{2}{*}{\textbf{\thead{Machine \\ Learning}}} & One-class Learning & Self-Organizing Map\citep{Kareth2022AD_HMM} \\
      \cline{2-3}
       & Multi-class Learning & Convolutional Neural Network\citep{li2021ADviaSOM} \\
      \hline
    \end{tabular}
\end{table}

\textbf{Statistical model based methods}: These supervised methods utilize non-parametric or parametric models to estimate the normal intensity distribution within defect images and then detect pixels out of the confidence interval as anomalies. For instance, Parzen Windowing method\citep{desforges1998Parzen} is utilized by Veracini et al.\citep{Veracini2011Nonparametric} to estimate the probability density function of the image background. Hierarchical model
\citep{Gao02092023} estimate the image background with the multivariate Gaussian model under a structural assumption of rapid mean changes and slow variance changes. Decomposition based methods \citep{Lahoti02102021,Wang04052022,MSEC2019} and Bayesian optimization \citep{Albahar1} have been investigated with parametric or nonparametric basis. Statistical model based methods can be effective in low-dimensional datasets but possess limitations in detecting small anomalies, such as minor scratches or grind marks, in real-world industrial scenarios with unstable environment like frequent-changing lighting due to their solid model assumptions. 

\textbf{Domain transformation based methods}: These methods transform image information from pixel domain to another space with less sensitivity to anomalies and noise to accomplish superior anomaly detection performance. An effective practice is to conduct a spatial-frequency domain transformation through various filters. In addition to conventional methods like Fourier transform\citep{tsai2003automated}, \cite{wilts2000auto} applied a multi-channel Gabor filter to extract texture features on steel surfaces; \cite{mehmood2010anomaly} utilized a 2D wavelet transform to filter infrared sensor images so that effective anomaly detection can be obtained on the reconstructed sub-band cube. Multiscale Geometric Analysis (MGA) methods, such as Ridgelet Transform\citep{emmanuel1999ridge}, Curvelet Transform \citep{starck2001curvelet}, and Contourlet Transform \citep{do2005contourlet}, have been introduced to handle vector anisotropy in high-dimensional data. 
In addition to transforming images into the frequency domain, methods that convert the spatial information into other special domains have been applied for anomaly detection. For instance, \cite{hma2006HoughTrans} applied a two-stage Hough Transform to extract complex spatial patterns from the background image without frequency features involved.
In general, domain transformation based methods exhibit fast computation speeds and reliable performance in scenarios with simple textures, strong periodicity, and high real-time requirements. However, their effectiveness can be limited when the periodicity embedded in images is influenced by translation, rotation, or other variations. Furthermore, it is concerned that domain transformation may cause information loss in the internal topological structure of original images.

\textbf{Clustering based methods}: These methods are generally unsupervised and can be classified into distance-based and density-based methods according to the mechanism. Distance-based methods such as K-nearest Neighbors, Local Outlier Factors, and Isolated Forest assume outliers away from normal data as anomalies. These methods could suffer from the curse of dimensionality when facing high-dimensional datasets. On the other hand, Density-based methods perform better in high-dimensional clustering tasks because they assume anomalies lie in regions with sparse distributed data. DBSCAN (density-based spatial clustering of applications with noise) methods have limitations when anomalies possess large differences in densities among dimensions. Efficient improvements include comprehensive sampling \citep{Tu2020HADusingDWD} and new kernel design for density estimation \citep{Hu2020localKDE}. Clustering based methods possess an advantage over other methods because there are no labels required, but they have limitations when textures exist in images, which can result in misidentification between textures and anomalies.
    
\textbf{Image reconstruction based methods}: These methods conduct data decomposition to project images into a low-dimensional subspace, where anomalies are separated from other components before being projected backward and reconstructed in the original space. 
Excellent example include Principal Component Analysis (PCA) and its variants\citep{yan2017anomaly,yan2018real}.
Such methods have decreased their dependence on pre-defined statistical models at a cost of high computation complexity in solving the corresponding optimization problem.
On the other hand, sparsity can be utilized as a distinguishable indicator of data decomposition process. For instance, sparse coding is integrated into neural blocks of AnomalyNet proposed by \cite{zhou2019anomalynet}
to help with efficient dictionary learning. Such methods have obtained higher generalization for complex scenarios like textured images, while introducing a high reliance on well-constructed datasets to learn a sufficient sparse dictionary. 
In conclusion, image reconstruction based methods possess concise model structures but usually take up larger space to store pre-learned feature information. 
    
\textbf{Machine learning based methods}: These methods utilize neural networks to automatically learn deep features from images to conduct anomaly detection\nocite{xu2025change}. They can be generally divided into one-class learning and multi-class learning methods. 
For instance, \cite{li2021ADviaSOM} projected defect images onto a 2D topological space using a Self-Organizing Map to conduct anomaly detection with only defect-free images as training data. \cite{cao2022semiknowledge} proposed a semi-supervised CNN method based on Knowledge Distillation, with defect-free images and a small set of defect images as training data, to enhance the ability to recognize small anomalies. However, obtaining sufficient defect images can be challenging in real-world industrial processes. \cite{tayeh2020distanceAD} applied Random Erasing on defect-free images to generate reliable defect samples. \cite{Zavrtanik2021reconstruction} trained an encoder-decoder network on defect-free samples and detected images with poor reconstruction effect as anomalies. To combine machine learning with conventional anomaly detection is also a universal attempt. \cite{Kareth2022AD_HMM} introduced the Hidden Markov Model (HMM) for temporal anomalies detection, while \cite{wyatt2022anoDDPM} optimized the Markov chain and Gaussian models within the traditional denoising diffusion probabilistic model (DDPM) to capture large image anomalies.
In general, machine learning based methods can achieve good anomaly detection performance with a well-balanced and sufficient training dataset. However, their applications can be limited by the scarcity of high-quality training data and anomaly labels, as well as requirements of robustness and reliability in real-world industrial scenarios.

\subsection{RPCA and SSD} 
With the primary focus of obtaining a deterministic estimation of image components, image reconstruction based methods are considered as priority alternatives for textured image anomaly detection. Specifically, analytical methods driven by data decomposition \citep{MSEC2019} including Principal Component Analysis (PCA) and Smooth-sparse Decomposition (SSD) are highlighted in this study. 

PCA is a classical method for reducing data dimensionality, which has found extensive applications in image retrieval, anomaly detection, and quality control. The rationale of PCA is to project high-dimensional data onto a lower-dimensional space that maximizes the variance among different principal components,
which enables subsequent data compression and feature extraction. Robust Principal Component Analysis (RPCA) and its expansion methods were proposed to enhance the robustness of PCA against outliers and improve its ability under high-dimensional scenarios. The original data is generally considered as a composition of low-rank normal data, sparse anomalies, and random noise in RPCA \citep{candes2011robust} and solved by convex optimization.
Whereas, RPCA methods still exhibit inefficiency when applied to large-scale and high-dimensional datasets in industrial scenarios, which has motivated other PCA-driven methods like Kernel-PCA and SSD.

Under the assumption of the smooth image background and sparse anomalies, \cite{yan2017anomaly} established the fundamental framework of SSD and expanded it temporally as the Spatio-Temporal Smooth Sparse Decomposition (ST-SSD,\citep{yan2018real}). Compared with other approaches for image anomaly detection, SSD and ST-SSD methods provided an integrated approach for denoising and anomaly detection. The simultaneous decomposition enhances the efficiency and reliability of anomaly detection in images with smooth backgrounds. 

However, these methods lack the necessary estimation of normal textures embedded in original images. 
\textcolor{black}{
For example, RPCA relies on the assumption of low-rank background and sparse anomalies to separate image components, which is challenging to eliminate texture related information and accomplish reliable performance.
On the other hand, SSD related methods can accomplish promising estimation of image components including background and anomalies with considerable reliance on the smoothness and sparsity. However, these two characteristics are insufficient to distinguish highly uncertain textures. It could be time-consuming for SSD related methods to construct appropriate mathematical basis for quasi-periodic textures, while a universal basis setting generally results in overestimating anomalies mixed with textures.
In summary, the applicability of these methods in textured image anomaly detection is limited due to deficiency in their core models, which is a primal focus of this paper.
}


{\color{black}{
\subsection{Proposed TBSD Method}

Targeted at efficient anomaly detection in textured images, our proposed Texture Basis Integrated Smooth Decomposition (TBSD) method can be classified into the \textbf{Image Reconstruction} based methods in Table \ref{table: liter_imageADmethods}. 
As shown in Table \ref{table: liter_methodCompare}, another two representative methods including RPCA and SSD are selected to compare with the proposed TBSD method. From the perspective of qualitative analysis, their main differences lie in the \textbf{involved components} and corresponding \textbf{basic assumptions}.

\begin{table}[h]
    \centering  
    \setlength{\tabcolsep}{3pt}
    \caption{\textcolor{black}{Descriptive comparison of Image Reconstruction based methods}}
    \label{table: liter_methodCompare}
    \textcolor{black}{
    \begin{tabular}{|c|c|c|c|c|}
      \hline
      \textbf{Method} & 
      \textbf{Low-Rank} & 
      \textbf{Smoothness} & 
      \textbf{Sparsity} & 
      \textbf{Texture} \\
      \hline
      \textbf{RPCA}\citep{candes2011robust}
      & Yes & No & Yes & No \\
      \hline
      \textbf{SSD}\citep{yan2017anomaly,yan2018real}
      & Yes & Yes & Yes & No \\  
      \hline
      \textbf{Proposed TBSD} 
      & Yes & Yes & Yes & Yes \\ 
      \hline
    \end{tabular}
    }
\end{table}

As previously mentioned, RPCA method has limitations in accurately separating image background from anomalies due to the insufficiency of low-rank assumption on background. SSD method solves that problem by introducing an additional smoothness assumption on image background, while this model may not be fully applicable to textured images, as sparse textures can easily be mistaken for anomalies.

When compared with those two Image Background base methods, our proposed TBSD method firstly takes the problem of texture misidentification into consideration. Subsequently, a rigor mathematical system of \textbf{quasi-periodicity} is formulated in Section \ref{s:perio} to provide another unique property for efficient separation of textures from anomalies. Quantitative modeling and detailed explanations could be found in Section \ref{s:method}.

The unique advantages and benefits of adopting the proposed TBSD method can be concluded as follows:
\begin{enumerate}[label=\arabic*.]
    \item \textbf{Significant expansion of data decomposition on various images}: The proposed TBSD method has expanded the application of data decomposition methods to textured images. It efficiently solves the texture misidentification problem in anomaly detection, which brings much more tolerance to industrial image processing.

    \item \textbf{Rigorous deterministic modeling method for image texture}: Quasi-periodicity is proposed with mathematical formulation for deterministic modeling of image textures. Leveraging this unique property enables efficient decomposition and reconstruction of textures, which in turn enhances the accuracy of anomaly detection in textured images.

    \item \textbf{Novel methodology of integrating prior knowledge into optimization}: On a basis of quasi-periodicity property, it is proposed that texture basis functions could be automatically learned in a form of sparse dictionary rather than manual construction. This creative approach enhances the universality and cost-effectiveness of image data decomposition methods. Both simulation (Section \ref{s:simuexp}) and real-world experiments (Section \ref{s:realexp}) have sufficiently validated its efficiency and practicability.
\end{enumerate}

}
}


\section{Quasi-periodicity of Texture Component} \label{s:perio}

\subsection{Decomposition and Examples of Source Image}
\label{subsec:mainComponents}
\textcolor{black}{
It is worth noting that the targeted images studied in this paper can be quasi-periodic textured images with smooth backgrounds and sparse anomalies. These textured images can be modeled as follows:
}

Suppose a source image $Y$ is represented by a matrix with the size of $m\times n$ in pixel. $Y$ can be decomposed into four distinct components as follows:
\begin{equation}
    Y=C_{bg}+C_{tex}+C_a+e
    \label{eq:decomposition}
\end{equation}
\renewcommand\labelenumi{(\theenumi)}
\begin{enumerate}
    \item \textbf{$C_{bg}$: Low-rank smooth background}: \newline
    \indent It represents the image background. This component should be both low-rank and smooth. A smooth basis is constructed to estimate it in data decomposition.
    \item \textbf{$C_{tex}$: High-rank quasi-periodic texture}: \newline
    \indent It represents a series of textures and should be treated as a normal component. It exhibits high-rankness because of the complexity and variability of textures and possesses quasi-periodicity as elements of those structures are similar in shape and size. A series of texture basis functions are specifically generated to extract this component. 
    \item \textbf{$C_a$: High-rank sparse anomaly}: \newline
    \indent It represents anomalies that are sparsely distributed within defect images and equal to \textbf{$\emptyset$} in a defect-free image. It exhibits sparsity because those anomalies are usually small in both sizes and total proportion compared to the entire image. 
    This component is estimated by residuals after data decomposition and denoising.
    \item \textbf{$e$: Random noise}: \newline
    \indent It represents random noise in the source image, usually from measurement precision or inherent variability. 
\end{enumerate}
\label{def:image component}

\begin{figure}[h]
    \centering
    \includegraphics[width=0.8\linewidth]{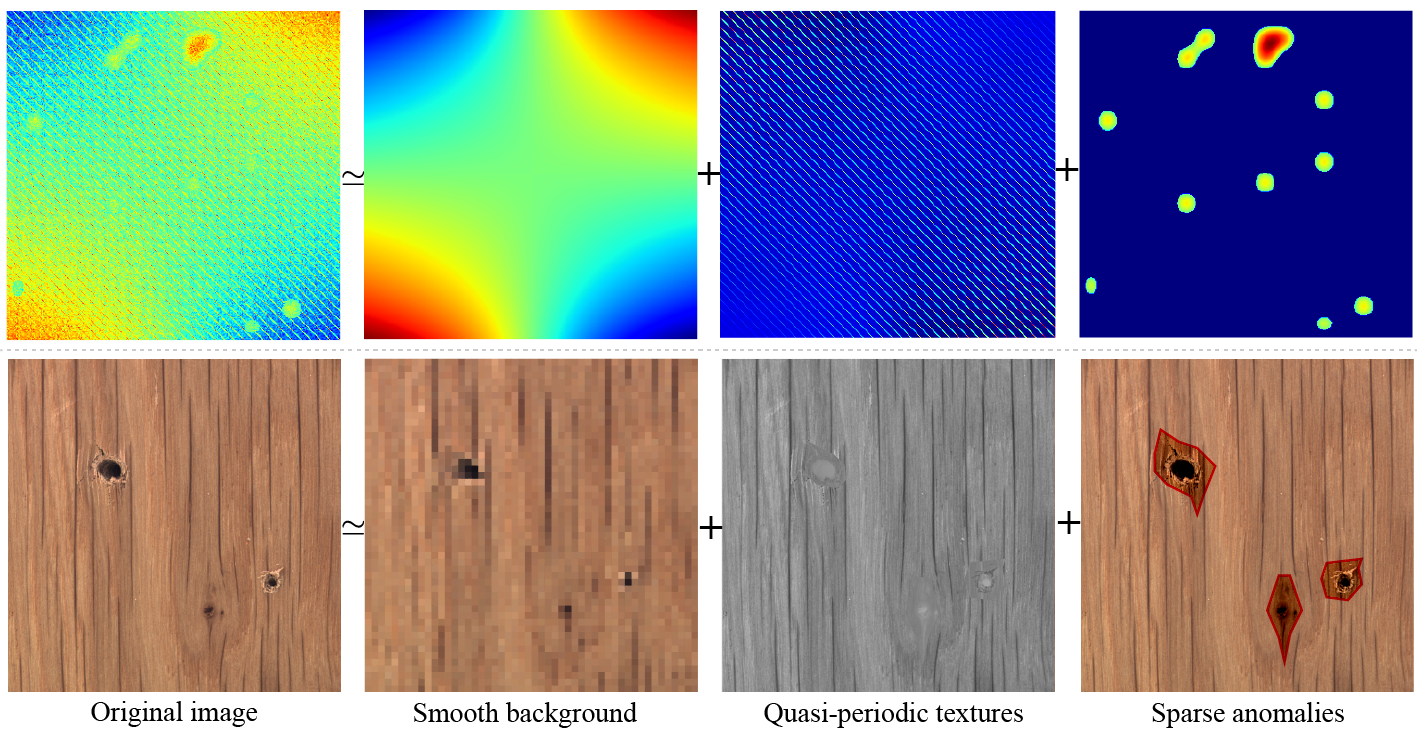}
    \caption{\textcolor{black}{Background, textures, and anomalies in simulation/real-world image examples}}
    \label{fig:image_decomposition}
\end{figure}

\begin{figure}[H]
    \centering
    \includegraphics[width=0.8\linewidth]{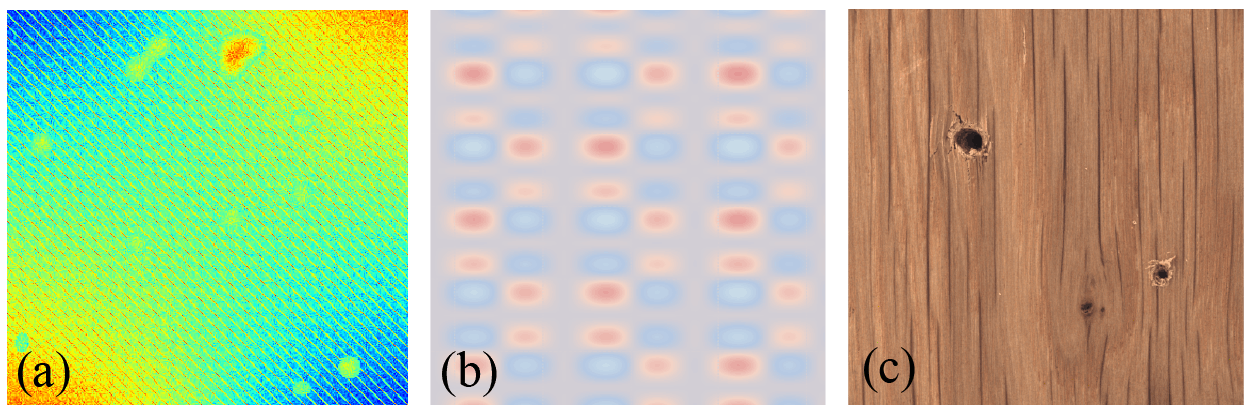}
    \caption{\textcolor{black}{Examples of textured images: (a) Manually constructed image with mainly linear textures for Simulation Study; (b) Manually constructed image with theoretically quasi-periodic textures; (c) Real-world image of wooden surfaces for Case Study}}
    \label{fig:textures_images}
\end{figure}

\textcolor{black}{
Figure \ref{fig:image_decomposition} 
is a typical example to illustrate how those characteristics including smoothness, quasi-periodicity, and sparsity are implied in corresponding components in real-world images.
Practically, there are considerable real-world images with seamlessly varying background, almost similar texture patterns, and sparse anomalies, such as Figure \ref{fig:textures_images}. These images lay a foundation for the generalizability of the textured image anomaly detection.
}

\textcolor{black}{
There is a considerable proportion of real-world textured images. However, those textures could generally possess complex and highly uncertain patterns while rarely having reliable manual annotations, so that it is essential and valuable to develop an unsupervised approach to efficiently learn associated features.}
\textcolor{black}{
Under this circumstances, \textbf{quasi-periodicity} is proposed as another unique property besides sparsity to separate image textures from anomalies. 
It is targeted at supporting efficient decomposition and reconstruction of image textures so as to enhance anomaly detection in textured images. 
}

\textcolor{black}{
This section would derive definitions and theorems about quasi-periodicity from universally adopted periodicity across different dimensions. Nomenclature is concluded in Table \ref{tab:s3notations}. The objective is to provide mathematical basis for extracting common patterns embedded in quasi-periodic textures like examples illustrated in Figure \ref{fig:textures_images}. 
}

\subsection{One-dimensional Periodicity} 
\label{s3subsec:1d_periodicity}
    \begin{definition}
        A set of color intensity $s$ sampled from a piece of image is said to be a \textbf{signal set} if that set is discrete, finite, and real-valued. 
        It follows that:
        \begin{enumerate}[leftmargin=2em, label=\roman*.]
            \item In the one-dimensional (1D) case, a 1D signal set is denoted as $S$ and consists of real-valued color intensity points (e.g., gray value of each pixel as $s$), which can be expressed as $S=\{s_1, s_2, s_3, \dots, s_n\}\in \mathbb{R}^1$; 
            
            \item In the two-dimensional (2D) case, a 2D signal set is denoted as $S^2$ and consists of multiple 1D signal sets, which can be expressed as $S^2=\{S_1, S_2, S_3, \dots, S_n\}\in \mathbb{R}^2$; 
            
            \item Particularly, a 2D signal set is said to be a \textbf{signal matrix} if all lengths of component 1D signal sets $\{S_j\vert S_j \in S^2\}$ are the same; A signal matrix can also be regarded as a sampled patch from the original image.
        \end{enumerate}
        \label{def:signal_set}
    \end{definition}

    \begin{definition}
        A 1D signal set $S$ is said to be \textbf{periodic} if $\exists T\in \mathbb{Z}^+, T \in (1, n)$, there exists $s_{i+T}=s_i$ for every $s_i\in S, \forall i \in [1, n-T]$.
        \label{def:1D_periodicity}
    \end{definition}
    
    As expressed in Definition \ref{def:1D_periodicity}, the periodicity with respect to a 1D signal set $S$ possesses three essential elements, which includes the definition domain $\mathbb{F}_n = \{i\in \mathbb{Z}^+ \vert 1\le i\le n\}$, the periodic length $T$, and the periodic mode $S_T = \{s_1, s_2, s_3, \dots, s_T\vert s_j\in S]\}$. 
    As shown in Figure \ref{fig:quasi_loosen_constraints}, a 1D periodic signal set $S$ can be regarded as a connected duplication of its periodic mode $S_T$ on equal-length segments.
    A 1D periodic signal set $S$ can be denoted with a $do(\cdot)$ operator as follows:
    \begin{equation}
    S=do(S_T, \mathbb{F}_n)
    \end{equation}

    Periodic 1D signal sets are commonly recognizable due to their distinctive characteristics, which is the same for their composition (Proposition \ref{proposition:composite_periodicity}). 

    \begin{theorem}
        A composite 1D signal set $S$, which is a linear combination of periodic 1D signal sets $\{S^{(k)}\vert S^{(k)}=do(S_{T^{(k)}}, \mathbb{F}_{n^{(k)}}), 1\le k \le K, k\in \mathbb{Z}^+\}$, is also periodic if and only if all ratios of any two periodic lengths $\{T^{(k_1)}/T^{(k_2)}\vert 1\le k_1, k_2\le K, k_1\neq k_2\}$ are rational. 
        \label{theorem:composite_periodicity_constraint}
    \end{theorem}

    \begin{proposition}
        For a composite 1D periodic signal set $S=\sum_{k=1}^{K}\beta^{(k)}S^{(k)}$, where $S^{(k)}=do(S_{T^{(k)}}, \mathbb{F}_{n^{(k)}})$ and $\beta^{(k)}\in \mathbb{R}$, it follows that $S=do(S_T, \mathbb{F}_N)$ and:
        \begin{enumerate}[leftmargin=2em, label=\roman*.]
            \item $S$ possesses a definition domain $\mathbb{F}_N$ where $N=max\{n^{(k)}\vert 1\le k\le K\}$;
            \item $S$ possesses a periodic length $T$ being the least common multiple of $\{T^{(k)}\vert 1\le k\le K\}$ (denoted as $\{T^{(k)}\vert 1\le k\le K\}^*$);
            \item $S$ possesses a periodic mode $S_T=[\sum_{k=1}^{K}\beta^{(k)}S_{T^{(k)}}]_{:T}$.
        \end{enumerate}
        \label{proposition:composite_periodicity}
    \end{proposition}

    Theorem \ref{theorem:composite_periodicity_constraint} has revealed the necessary and sufficient condition of creating another 1D periodic signal set by linearly combining 1D periodic signal sets. That condition can be easy to accomplish in real-world practices with all signals constrained to be rational.

\subsection{Two-dimensional Periodicity} \label{s3subsec:2d_periodicity}
    \begin{definition}
        A 2D periodic signal set $S^2$ is said to be \textbf{periodic} if $S^2 = S^{[1]} \otimes S^{[2]}$, where $S^{[1]}, S^{[2]}$ are composite 1D periodic signal sets with periodic lengths being $T^{[1]}$ and $T^{[2]}$, respectively. It follows that $S^2$ possesses a definition domain $\mathbb{F}^2_{N^{[1]}, N^{[2]}}$, a periodic size $T^2 = (T^{[1]}, T^{[2]})$, and a periodic mode $S^2_{T^2} = [S^2]_{(:T_1, :T_2)}$.
        \label{def:2D_periodicity}
    \end{definition}
    According to Definition \ref{def:2D_periodicity}, it is possible to obtain a 2D periodic signal set by conducting outer production on two composite 1D periodic signal sets. Each row of such a 2D periodic signal set $S^2$ can be regarded as the composite 1D periodic signal set $S^{[2]}$ weighted by elements from $S^{[1]}$, which is similar for each $S^2$'s column.
    
   As illustrated in Figure \ref{fig:1Dto2D}(a), 2D signal sets like images with periodicity are common in the real world as that characteristic is relatively stable across dimensions.
   It is effective to directly extract common modes from periodic images, while doing the same could be difficult when conducted on images with quasi-periodicity.
\subsection{One-dimensional Quasi-periodicity} \label{s3subsec:1d_quasi_periodicity}
    \begin{figure}[h]
        \centering
        \includegraphics[width=\linewidth]{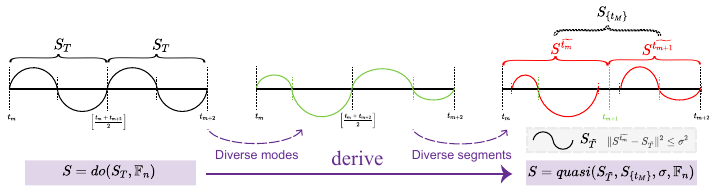}
        \caption{\textcolor{black}{How quasi-periodicity derives from periodicity: an originally periodic 1D signal set $S$, with uncertainty attached onto lengths and elemental sets of its periodic segments, could be transformed into a 1D quasi-periodic signal set. (For the convenience of illustration, discrete signal sets are connected as continuous curves in this figure).}}
        \label{fig:quasi_loosen_constraints}
    \end{figure}

    \vspace{-1em}
    \begin{definition}
        A 1D signal set $S$ is said to be \textbf{quasi-periodic} if there exists a segmentation $S_{\{t_M\}}$ where $\exists M \in \mathbb{Z}^+ \setminus{1}$, a quasi-periodic length $T\in (1, \mathop{\min}_{m}\{t_m\}]$, and a quasi-periodic mode $S_{\tilde{T}}$ so that $\exists \sigma > 0, 0\le {\|S^{\widetilde{t_m}}-S_{\tilde{T}} }\|^2\le \sigma^2$ is satisfied for every $S^{\widetilde{t_m}} \in S_{\{t_M\}}$.
        \label{def:1D_quasi_periodicity}
    \end{definition}
    \vspace{-2em}
    
    As expressed in Definition \ref{def:1D_quasi_periodicity}, the quasi-periodicity with respect to a 1D signal set $S$ possesses five essential elements, which includes the definition domain $\mathbb{F}_n = \{i\in \mathbb{Z}^+ \vert 1\le i\le n\}$, the quasi-periodic segmentation $S_{\{t_M\}}$, the quasi-periodic length $T$, the quasi-periodic mode $S_{\tilde{T}}$, and the quasi-periodic control limit $\sigma$. A 1D quasi-periodic signal set $S$ can be denoted with a $quasi(\cdot)$ operator as follows:
    \begin{equation}
    S=quasi(S_{\tilde{T}}, S_{\{t_M\}}, \sigma, \mathbb{F}_n)
    \end{equation}

    The definition of quasi-periodicity is an extension of that of periodicity. As illustrated in Figure \ref{fig:quasi_loosen_constraints}, it has loosened constraints on strict equality among lengths and corresponding elemental sets of periodic segments defined in Definition \ref{def:1D_periodicity}.

    \begin{definition}
        A composition of 1D quasi-periodic signal sets can be expressed as $S=\sum_{k=1}^{K}\beta^{(k)}S^{(k)}$, where $S^{(k)}=quasi(S_{\widetilde{T^{(k)}}}, S_{\{t_M^{(k)}\}}, \sigma^{(k)}, \mathbb{F}_{n^{(k)}})$ and $\beta^{(k)}\in \mathbb{R}$.
        \label{def:composite_quasi_periodicity}
    \end{definition}

    \begin{theorem}
        A composition $S$ of 1D quasi-periodic signal sets, as defined in Definition \ref{def:composite_quasi_periodicity}, is said to be quasi-periodic if there exists $\sigma>0$ so that $(\sum_{k=1}^{K}{\beta^{(k)}}^2) [\sum_{k=1}^{K}(\| {(S^{(k)})}^{\widetilde{t_m}} \|_2 + \sigma^{(k)})^2]
        \le \sigma^2$ is satisfied for every segment $\widetilde{t_m}$ where all components $S^{(k)}$ show quasi-periodicity. 
        \label{theorem:composite_quasi_periodicity_constraint}
    \end{theorem}

    Similar with the periodic case (Proposition \ref{proposition:composite_periodicity}), a linear composition of 1D quasi-periodic signal sets can be expressed in Definition \ref{def:composite_quasi_periodicity}. 
    However, quasi-periodicity of a composite 1D signal set tends to be weakened as number and diversity of its components increases. Theorem \ref{theorem:composite_quasi_periodicity_constraint}, proved in Appendix \ref{appsec:Cauchyproof}, has shown that both increase of component number $K$ and diversity among quasi-periodic segmentation $S_{\{t_M^{(k)}\}}$ reduce the possibility and scale of all components $S^{(k)}$ share a quasi-periodic segment $\widetilde{t_m}$. Though this phenomenon makes the constraint easier to be satisfied as $\widetilde{t_m}$ becomes rare and small, the quasi-periodicity of $S$ can be weakened with the complexity of components increases.

\subsection{Two-dimensional Quasi-periodicity} \label{s3subsec:2d_quasi_periodicity}
\begin{figure}[H]
    \centering
    \includegraphics[width=0.8\linewidth]{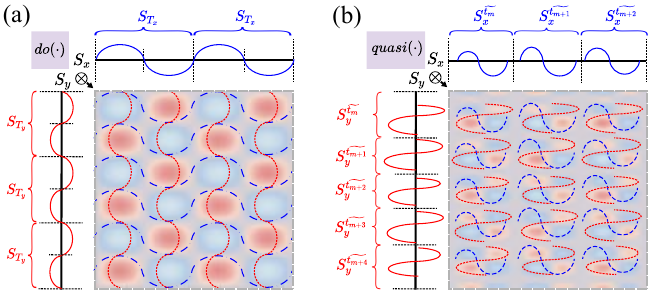}
    \caption{\textcolor{black}{Producing 2D matrix with (a)periodic /(b)quasi-periodic 1D signal sets by outer product operation: (a) still possess evident periodicity which can be directly captured; (b) consists of sparse textures with similar patterns but high uncertainty
    }}
    \label{fig:1Dto2D}
\end{figure}

The same outer product operation as Definition \ref{def:2D_periodicity} can be applied to generate a 2D signal set from a 1D quasi-periodic signal set as illustrated in Figure \ref{fig:1Dto2D}(b). 
Such a constructed 2D signal set is an expected image with quasi-periodicity. Whereas, 2D quasi-periodic image may either show globally disordered patterns or possess too small quasi-periodic segments to be recognized.
It could be inefficient to directly capture common modes in such an image, while its 1D component quasi-periodic signal sets are much easier to be modeled.

Therefore, for the texture component in an arbitrary textured image, which is considered as a 2D quasi-periodic signal set, it is suggested to capture common texture patterns after reducing dimension for effectiveness and efficiency. 
Practically, the dimensionality is reduced along a $\theta$ direction in the polar coordinate system, with the image center as the pole and the horizontal right as the polar axis. It follows that if the gained 1D signal set is quasi-periodic, then $S^2$ is denoted to possess quasi-periodicity along the $(\theta+\pi/2)$ direction, which is orthogonal to the dimension-reduced direction $\theta$. 

\subsection{Section Conclusion}
\label{s3subsec:conclusion}
The proposed quasi-periodicity theoretically derives from periodicity (Definition \ref{def:1D_periodicity}). The necessity and significance of introducing quasi-periodicity into anomaly detection on textured images can be concluded as follows:
\begin{enumerate}
    \item Intuitively, the core idea behind quasi-periodicity is to summarize features of those ``similar but not identical" texture patterns existing in textured images.
    
    \item Practically, quasi-periodicity is introduced to describe and model the potentially common texture patterns embedded in textured images.

    \item Mathematically, with an assumption of textures possess quasi-periodicity, it is proved to be more efficient to extract common texture patterns after dimensional reduction. 
\end{enumerate}

In addition, when it comes to periodic scenarios, 
the quasi-periodicity
would shrink into a concise model as defined in Definition \ref{def:1D_periodicity}, \ref{def:2D_periodicity}. Related models and methods could be utilized to extract periodic modes and reconstruct similar textures in other images.


\section{Texture Basis Integrated Smooth Decomposition \\ (TBSD) Method} \label{s:method}
In this section, the Texture Basis Integrated Smooth Decomposition (TBSD) methodology is proposed (Subsection \ref{subsec:optimModel}). It consists of two crucial processes including (\Rmnum{1}) texture basis function learning and (\Rmnum{2}) anomaly detection. 
With a smooth basis $B$ pre-constructed in Subsection \ref{subsec:smooth_basis}, Process \Rmnum{1} (Subsection \ref{subsec:BtLearnProcess}) learns a set of texture basis functions $B_t$ from textures $\widehat{C_{tex}}$ in a defect-free image $Y$. Then Process \Rmnum{2} (Subsection \ref{subsec:ADprocess}) utilizes both $B$ and $B_t$ to targetedly separate anomalies $\widehat{C_a}$ in a defect image $\dot{Y}$. 
Algorithms to conduct TBSD are further developed based on optimization models in Table \ref{tab:optimModel}.

\vspace{-2em}
\subsection{Proposed Optimization Model}
\label{subsec:optimModel}
The proposed TBSD method could be concluded into two mathematical optimization models as separately constructed in Table \ref{tab:optimModel} with notations in Table \ref{tab:s4notations}.

Model \Rmnum{1}, corresponding to the texture basis function learning process, is targeted at learning a set of texture basis functions $B_t$ from a defect-free image $Y$, along with a pre-constructed smooth basis $B$ as input. This model consists of two phases. Phase \Rmnum{1}-\rmnum{1}\: decomposes $Y$ to estimate textures $\widehat{C_{tex}}$ following Algorithm \ref{algorithm:LowRankDecomposite}; Then Phase \Rmnum{1}-\rmnum{2}\: learns $B_t$ along different directions $D$ where $\widehat{C_{tex}}$ possesses quasi-periodicity. Algorithm \ref{algorithm:QuasiPeriodicDetect} and \ref{algorithm:RotateSampling} are proposed to find an acceptable solution for $B_t$.

Subsequently, Model \Rmnum{2}, corresponding to the anomaly detection process, would be conducted on a defect image $\dot{Y}$ following Algorithm \ref{algorithm:DefectImageDecomposition}. The pre-learned texture basis $B_t$ serves as a sparse dictionary for reconstructing textures $\widehat{C_{tex}}$ in $\dot{Y}$, so that misidentification problem in anomaly detection can be targetedly prevented.

\begin{table}[H]
    \centering
    \caption{Optimization Model for defect-free / defect images}
    \resizebox{\textwidth}{!}{
    {\color{black}
    \begin{tabular}{c|c|c|c}
    \toprule
         & \multicolumn{2}{c}{\textbf{(\Rmnum{1}) defect-free image}} \vline  & \textbf{(\Rmnum{2}) defect image} \\
         \midrule
        \textbf{Input} 
        & \textbf{(\Rmnum{1}-\rmnum{1})}\quad $Y, B$
        & \textbf{(\Rmnum{1}-\rmnum{2})}\quad $\widehat{C_{tex}}$
        & $\dot{Y}, B, B_t$
        \\
         \midrule
        \textbf{Objective} 
        & $\underset{\theta}{\operatorname{argmin}} \; \|e\|^2+\lambda \theta^T R\theta+\gamma\Vert \widehat{C_{tex}}\Vert_1$ 
        & $\underset{D}{\operatorname{argmin}} \|\widehat{C_{tex}} - \widetilde{C_{tex}}\|^2$
        & $\underset{\theta, \theta_t}{\operatorname{argmin}} \; \|e\|^2+\lambda \theta^T R\theta +\gamma\|\theta_t\|_1 +\eta\Vert \widehat{C_a}\Vert_1$
        \\
        \textbf{Subject to:}
        &
        $\begin{matrix}
            \quad Y =\widehat{C_{bg}}+\widehat{C_{tex}} + C_a +e \\
            \widehat{C_{bg}} = B\theta, \: C_a = \emptyset \\
        \end{matrix}
        $
        & 
        $\begin{matrix}
            \widehat{C_{tex}} = \sum\limits_{d\in D}{\widehat{C_{tex}^{(d)}}}, \\
            \widetilde{C_{tex}} = \sum\limits_{d\in D}{B_t^{(d)}}
        \end{matrix}$
        &
        $\begin{matrix}
            \quad \dot{Y}=\widehat{\dot{C}_{bg}}+\widehat{\dot{C}_{tex}}+\widehat{\dot{C}_a}+e \\
            \widehat{\dot{C}_{bg}}=B\theta, \: \widehat{\dot{C}_{tex}}=B_t\theta_t \\ 
        \end{matrix}$ \\
        \midrule
        \textbf{Output} 
        & $\hat{\theta}$ 
        & $D$
        & $\hat{\theta}, \hat{\theta_t}$ \\
        \midrule
        \textbf{Estimator} & 
        $\begin{matrix}
        \widehat{C_{tex}} = S_{\gamma/2}(Y - B\hat{\theta})
        \end{matrix}$
        & 
        $B_t=\{B_t^{(d)}\vert d\in D\}$
        & $\widehat{C_{a}} = S_{\eta/2}(\dot{Y} - B\hat{\theta} - B_t\hat{\theta_t})$ \\        \bottomrule
    \end{tabular}
    }
    }
    \label{tab:optimModel}
\end{table}
In Table \ref{tab:optimModel}, $S_{\gamma}(x) = sgn(x) \cdot (\vert x\vert - \gamma)_{+}$, 
    $x_{+} = max(x, 0), \: sgn(x) = \left\{\begin{aligned}
        1, x>0 \\
        0, x=0 \\
        -1, x<0
    \end{aligned}\right.$
\newline

$\theta$ and $\theta_t$ are two core parameters to be estimated in the optimization model.
These two estimators can be solved by ridge regression utilizing the idea of Block Coordinate Descent (BCD) method [\cite{yan2017anomaly}], with related proofs in Appendix \ref{appsec:ridgeReg_TBFLP}, \ref{appsec:ridgeReg_ADP} and results shown as follows:
\begin{flalign}
    & 
    \hat{\theta} = \left\{
    \begin{aligned}
        & (B^T B+\lambda R)^{-1} B^T\cdot(Y-\widehat{C_{tex}}), \qquad \quad \:\: \text{for defect-free images $Y$}
        \\
        & (B^TB+\lambda R)^{-1} B^T\cdot(\dot{Y}-B_t\hat{\theta_t}-\widehat{C_a}), \quad \text{for defect images $\dot{Y}$}
    \end{aligned}\right.
    \\
    &  \hat{\theta_t} \approx B_t^T \cdot (Y-B\hat{\theta}-\widehat{C_a}) - \frac{\gamma}{2} sgn(B_t^T \cdot(Y-B\hat{\theta}-\widehat{C_a}))
\end{flalign}

\textcolor{black}{
It is worth noting that the quasi-periodicity proposed in Section \ref{s:perio} for image textures plays a critical role in the whole optimization model.
This assumption is firstly explicitly expressed as Phase \Rmnum{1}-\rmnum{2} \: in TBSD's optimization model. It regards textures $C_{tex}$ as a composition of quasi-periodic sub-textures $C_{tex}^{(d)}$ along different directions $d\in D$ and decomposes $\widehat{C_{tex}}$ extracted from defect-free images to construct a series of texture bases $B_t=\sum_{d\in D}{B_t^{(d)}}$.
Subsequently, this assumption is utilized in Phase \Rmnum{2} for estimation and reconstruction of textures in defect images, so as to separate them for anomaly detection.
In the technical application of Algorithm B.1, B.2, and B.3, this assumption ensures the targeted texture basis functions $B_t$ are \textit{learnable} and could be \textit{simplified}. \textit{Learnability} allows image textures to be represented and estimated by linear combinations of $B_t$. \textit{Simplifiability} allows sufficient construction of $B_t$ based on a small batch of defect-free images. Detailed explanations could be found in the subsequent Subsection \ref{subsec:BtLearnProcess}
}




\subsection{Estimation of Low-rank Smooth Background}
\label{subsec:smooth_basis}
{\color{black}{
There are two assumptions for the background component $C_{bg}$ in a source image $Y_{m\times n}$:
\begin{enumerate}
    \item \textbf{Low-rank assumption}: Background possesses significantly lower rank than other components:
    \begin{align}
        rank(C_{bg}) \ll rank(Y)
        & & rank(C_{bg}) \ll rank(Y-C_{bg})
        \label{eq:lowrank}
    \end{align}
    This assumption is crucial for separating the background from the original image. It ensures that the decomposition based on low rank in Algorithm 4.1 is conducted in a reasonable manner. It is illustrated in Figure \ref{fig:method_Bgenerate} that a limited set of basis functions is sufficient to estimate the image background $C_{bg}$ by leveraging this assumption.
    
    \item \textbf{Smoothness assumption}: Denote the intensity of pixel $P_{(i, j)}$ as $G_{(i, j)}$ and its neighborhood as $H(P_{(i, j)})$, the adjacent intensity difference $D$ should be constrained within a threshold value $\rho>0$:
    \begin{equation}
        \begin{aligned}
        D_{(i_2, j_2 )}^{(i_1, j_1 )} = \vert G_{(i_1, j_1 )}-G_{(i_2, j_2 )} \vert\le \rho \cdot (\vert i_1-i_2 \vert + \vert j_1-j_2 \vert)
        &,
        \\
        \quad P_{(i_2, j_2 )} \in H(P_{(i_1, j_1 )}) ,
        \:
        1\le i_1, i_2\le m, \: 1\le j_1, j_2\le n& 
        \end{aligned}
        \label{eq:smoothness}
    \end{equation}
    This assumption essentially restricts the intensity function across the pixel domain to be $\rho$-Lipschitz continuous and intuitively smooth. Consequently, smooth basis functions could efficiently approximate such an intensity variation to estimate a smooth image background.
\end{enumerate}

Under these assumptions, the low-rank smooth background component $C_{bg}$ can be generated with a set of smooth bases $B$ as shown in Figure \ref{fig:method_Bgenerate}. The smooth basis function is constructed using a L-degree B-spline function
with $L=3$ and smoothness $k=2$. With the corresponding coefficient matrix denoted as $\theta$, there exists: 
\begin{equation}
    \widehat{C_{bg}}=B\theta =B_y \theta _y (B_x \theta _x )^T=B_y \theta _y \theta _x^T B_x^T
    \label{eq:smoothnessdecom2}
\end{equation}

\begin{figure}[H]
    \centering
    \includegraphics[width=\textwidth]{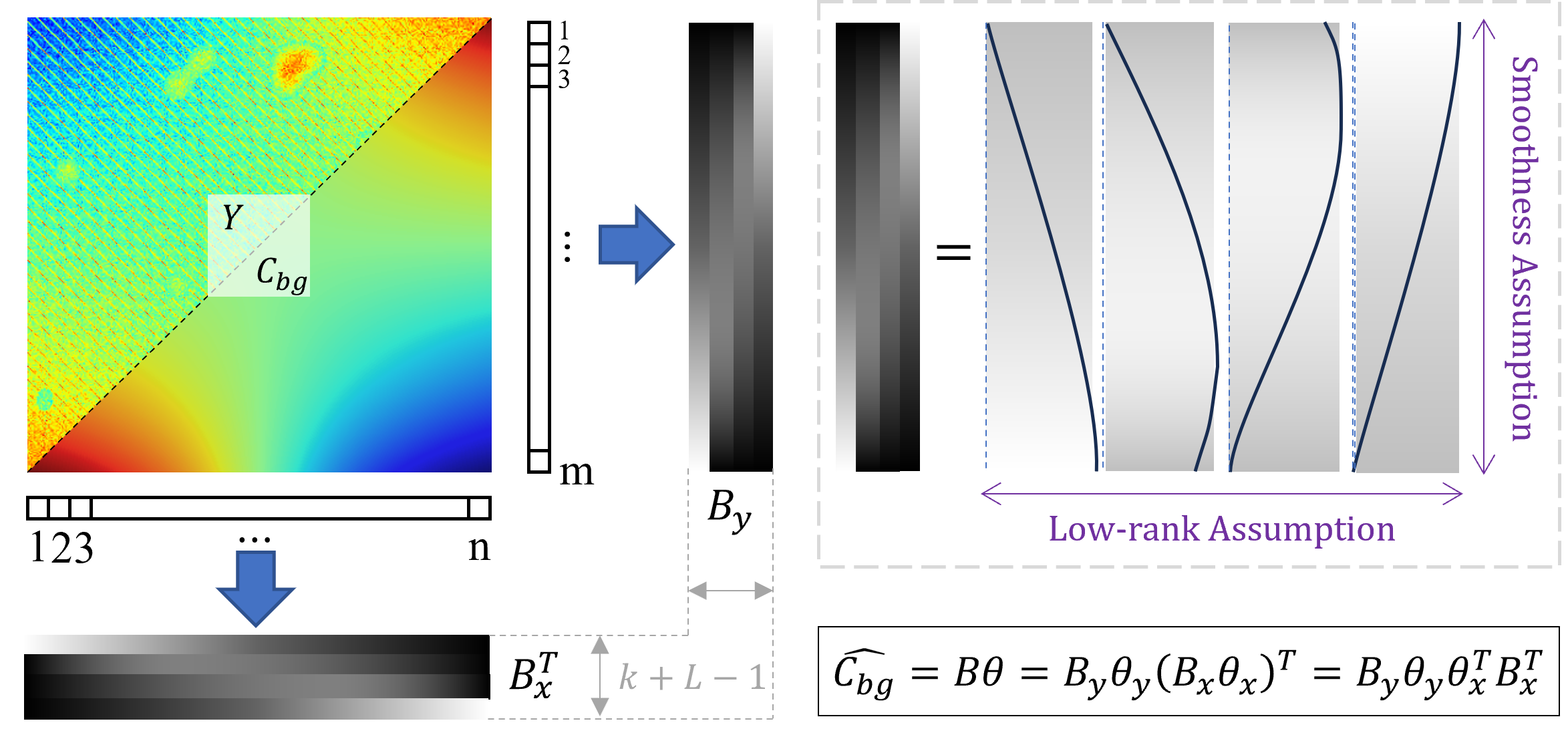}
    \caption{\textcolor{black}{Smooth Basis Function Generation based on 3-degree B-spline Curve}}
    \label{fig:method_Bgenerate}
\end{figure}
}
}
\vspace{-2em}

\subsection{Texture Basis Function Learning (TBFL) Process}
\label{subsec:BtLearnProcess}
\textcolor{black}{
The TBFL process (Table \ref{tab:optimModel}: Model \Rmnum{1}) conducts data decomposition on defect-free images. It is targeted at learning a set of texture basis functions that can be utilized in the next-step anomaly detection process.
In this first part (Model \Rmnum{1}-\rmnum{1}) of the training phase, Algorithm \ref{algorithm:LowRankDecomposite} is applied onto a defect-free image $Y$ to estimate embedding textures $\widehat{C_{tex}}$. This algorithm functions based on the rank difference between low-rank image background and other high-rank components. It estimates the smooth coefficient $\theta$ which reconstructs background $\widehat{C_{bg}}$ from smooth bases $B$, so that textures $\widehat{C_{tex}}$ could be separated.
}

\begin{algorithm}[h]
    \caption{Low-rank Data Decomposition for Defect-free Images} 
    \label{algorithm:LowRankDecomposite}
    \begin{algorithmic}[1] 
        \Require Source image data: $Y$, Smooth basis function: $B$, Roughness matrix:  $R$, Maximum iteration times:  $iterTimes$, Parameters: $\lambda, \gamma$  
        \Ensure Smooth background component: $\widehat{C_{bg}}$, High-rank texture component: $\widehat{C_{tex}}$
        \Function {LowRankDecomposite}{$Y, B, R, iterTimes, \lambda, \gamma$}  
        \State $\widehat{C_{tex}} \gets \emptyset$  
        \State $iIter \gets 0$
        \State $H \gets B (B^T B + \lambda R)^{-1} B^T$
        \While {$iIter < iterTimes$}  
        \State $\widehat{C_{bg}} \gets H (Y - \widehat{C_{tex}})$  
        \State $\widehat{C_{tex}} \gets S_{\gamma/2} (Y - \widehat{C_{bg}})$  
        \State $iIter \gets iIter + 1$
        \EndWhile  
        \State \Return{$\widehat{C_{bg}}, \widehat{C_{tex}}$}  
        \EndFunction   
    \end{algorithmic}  
\end{algorithm}

\textcolor{black}{
Model \Rmnum{1}-\rmnum{2}\: follows quasi-periodicity related properties specifically discussed in Section 3.5. It is solved to accomplish an appropriate texture basis $B_t$ based on the aforementioned defect-free textures $\widehat{C_{tex}}$. An ahead dimensional reduction is suggested to respectively extract 1D quasi-periodic modes from $\widehat{C_{tex}}$ instead of directly exploring in the original 2D case. Here $B_t$ serves as an overcomplete ``dictionary'' of texture patterns $\widehat{C_{tex}^{(d)}}$ along different directions $d\in D$ where $\widehat{C_{tex}}$ possesses 1D quasi-periodicity.
}
\begin{equation}
    \widehat{C_{tex}}=\sum_{d\in D} \widehat{C_{tex}^{(d)}},
\end{equation}
\vspace{-1.5em}

\textcolor{black}{
Technically, Algorithm \ref{algorithm:RotateSampling} and \ref{algorithm:QuasiPeriodicDetect} are proposed to determine $D$ and $\{\widehat{C_{tex}^{(d)}}\vert d\in D\}$ based on $\widehat{C_{tex}}$. Then Algorithm \ref{algorithm:KNBN} would generate a texture basis $B_t$ by density-based clustering.
For convenience, the $d$ and $(d+\pi/2)$ directions (corresponds to $\theta+\pi/2$ and $\theta$ direction in Section \ref{s3subsec:2d_quasi_periodicity}) are further denoted as the \textbf{texture expansion direction} and \textbf{texture extension direction}, respectively.
}

\textcolor{black}{
As complementary, Model \Rmnum{1}-\rmnum{2}\: is initially targeted at an optimal solution of $D$. However, texture bases $B_t$ would provide sparse elements to the next-step anomaly detection process for texture reconstruction, so a feasible solution for Model \Rmnum{1}-\rmnum{2}\: containing essential texture patterns is also acceptable in practice. 
}

\begin{figure}[h]
    \centering
    \includegraphics[width=0.8\textwidth]{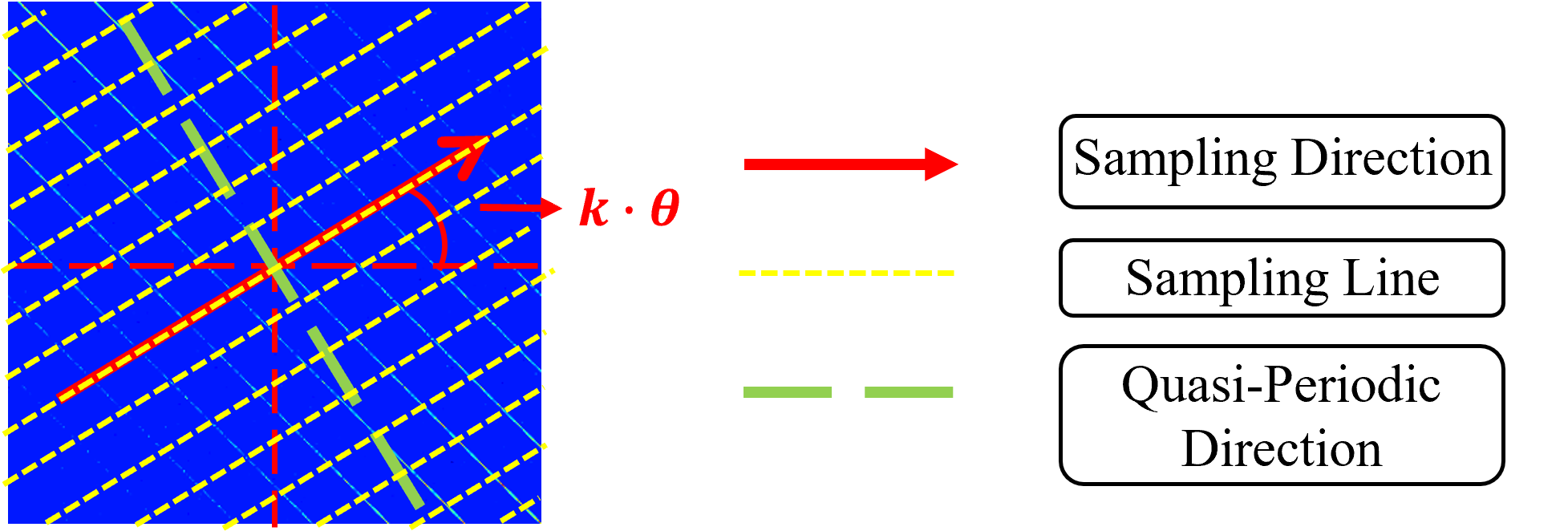}
    \caption{Schematic diagram of LSERA algorithm (Rotate times: $k$, Rotation angle: $\theta$)}
    \label{fig:method_LSERA}
\end{figure}

As Figure \ref{fig:method_LSERA} shows, Algorithm \ref{algorithm:RotateSampling}, denoted as Linear Sampling in Equal Rotation Angle (LSERA), results in a 1D signal set $\{S^{(k\theta)}\}$.
With a pre-defined threshold value $\phi_S>0$, the set of texture expansion directions $D$ is detected by the following criteria: 
\begin{equation}
    F(k\theta) = std(S^{(k\theta)})-std(S^{(k\theta+\pi/2)})
, \quad
    \left\{
    \begin{aligned}
        (k\theta+\pi/2) &\in D, \text{if } F(k\theta) > \phi_S \\
        (k\theta+\pi/2) &\notin D, \text{o.w.}
    \end{aligned}
    \right .
    \label{eq: quasiDetect}
\end{equation}

This criteria is the core of Algorithm \ref{algorithm:QuasiPeriodicDetect}: When the sampling direction aligns with the texture extension direction $\theta^*$, $S^{(k\theta)}$ exhibits a pattern of intertwined background and texture, resulting in a high standard deviation due to significant intensity difference between those two components, while the range of standard deviation in $S^{(k\theta+\pi/2)}$ would be low.

As supplementary, 
Algorithm \ref{algorithm:RotateSampling} theoretically root in the relative simplicity of detecting one-dimensional quasi-periodicity than that in the two-dimensional case (Section \ref{s3subsec:2d_quasi_periodicity}). The basic idea of this algorithm, which rotates a linear sampling line around a fixed point within images so that dimension-reduced information could be accomplished along specific directions. 


Here follows another practical operation for more efficient $B_t$ learning. According to the denotation in Section \ref{s3subsec:1d_quasi_periodicity}, $C_{tex}^{(d)}$ can be effectively expressed as a composition of a set of quasi-periodic texture modes $S_{\tilde{T}}^{(d)}$ along the texture expansion direction $d$, with the corresponding quasi-periodic length, segmentation, and control limit denoted as $T$, $S_{\{t_M\}}^{(d)}$, and $\sigma^{(d)}$, respectively. If there exists a spatial length $n$ for texture expansion along the $d$ direction, then $\{C_{tex}^{(d)}\}$ can be denoted as follows:
\begin{equation}
    C_{tex}^{(d)}=quasi(S_{\tilde{T}}^{(d)}, S_{\{t_M\}}^{(d)}, \sigma^{(d)}, \mathbb{F}_n)
    \label{eq:quasi_C_tex}
\end{equation}

Eq. \ref{eq:quasi_C_tex} has revealed that the primary learning basis of TBFL process can be further narrowed down from $\{\widehat{C_{tex}^{(d)}}\vert d\in D\}$ to $\{S_{\tilde{T}}^{(d)}\vert d\in D\}$. It is reasonable for quasi-periodic modes $\{S_{\tilde{T}}^{(d)}\vert d\in D\}$ to be sufficiently contained within partial areas of the original image $Y$, so segmentation could be conducted on $Y$ to learn effective texture basis $B_t$ based on image patches without significant impact on the overcompleteness of $B_t$. 

In conclusion, TBFL process is targeted at solving Model \Rmnum{1}-\rmnum{1}, \Rmnum{1}-\rmnum{2}\: in Table \ref{tab:optimModel} to accomplish an overcomplete set of texture basis function $B_t$
, which would be utilized in the subsequent anomaly detection process for texture reconstruction.

\subsection{Anomaly Detection Process}\label{subsec:ADprocess}

With the pre-learned texture basis functions $B_t$ from the TBFL process, Model \Rmnum{2} in Table \ref{tab:optimModel} could be solved for an optimal estimation of anomalies $C_a$ in defect images $\dot{Y}$. Algorithm \ref{algorithm:DefectImageDecomposition} is proposed for this anomaly detection process.
\textcolor{black}{
It is actually an extension of Algorithm \ref{algorithm:LowRankDecomposite} when the decomposed image $\dot{Y}$ contains anomalies $C_a\neq \emptyset$. This algorithm firstly separate background based on rank difference, then separate textures based on the proposed quasi-periodicity assumption, so as to efficiently detect image anomalies. Two core parameters including smooth coefficient $\theta$ and texture coefficient $\theta_t$ are solved as shown in Appendix C to iteratively improve the estimation of $\widehat{\dot{C}_{bg}}$ and $\widehat{\dot{C}_{tex}}$ for the final $\widehat{\dot{C}_a}$ detection.
}

\vspace{-2em}
\textcolor{black}{
It should be pointed out that removing image background by low rank based decomposition is necessary before operating on textures.
A defect-free image $Y$ is considered as a composite 2D signal set which consists of low-rank smooth background $C_{bg}$ and high-rank quasi-periodic textures $C_{tex}$. 
It is reasonably assumed that $Y$ is non-quasi-periodic 
because an additionally non-quasi-periodic background $C_{bg}$ can obscure the originally weak quasi-periodicity within 2D quasi-periodic signal set $C_{tex}$ (Section 3.5). Therefore, the texture basis functions $B_t$ should be learned individually using the quasi-periodic textures $C_{tex}$ in Model \Rmnum{1}-\rmnum{1}, as is the case for texture reconstruction in Model \Rmnum{2}.
}

\begin{algorithm}[H]
    \caption{Anomaly Detection based on Defect Images}
    \label{algorithm:DefectImageDecomposition}
    \begin{algorithmic}[1] 
        \Require Source image data:$\dot{Y}$, Smooth basis function:$B$, Texture basis function:$B_t$, Roughness matrix:$R$, Maximum iteration times:$iterTimes$, Parameters:$\lambda, \gamma, \eta, \phi_{B/T}$  
        \Ensure Smooth background component:$\widehat{C_{bg}}$, High-rank quasi-periodic texture component:$\widehat{C_{tex}}$, Sparse anomaly component:$\widehat{C_a}$
        \Function {DefectDecomposition}{$\dot{Y}, B, B_t, R, iterTimes, \lambda, \gamma, \eta, \phi_{B/T}$}  
        \State $\hat{\theta_t} \gets \emptyset$
        \State $\widehat{C_{a}} \gets \emptyset$  
        \State $iIter \gets 0$
        \State $H \gets B (B^T B + \lambda R)^{-1} B^T$
        \While {$iIter < iterTimes$}  
        \State $\widehat{C_{bg}} \gets H (\dot{Y} - B_t \hat{\theta_t} - \widehat{C_a})$  
        \State $\hat{\theta_t} \gets S_{\gamma/2} (B_t^T (\dot{Y} - \widehat{C_{bg}} -\widehat{C_a}) ) $
        \State $\widehat{C_{tex}} \gets B_t \hat{\theta_t}$  
        \State $\widehat{C_a} \gets S_{\eta/2} (\dot{Y} - \widehat{C_{bg}} - \phi_{B/T} \cdot \widehat{C_{tex}}) $
        \State $iIter \gets iIter + 1$
        \EndWhile  
        \State \Return{$\widehat{C_{bg}}, \widehat{C_{tex}}, \widehat{C_a}$}  
        \EndFunction   
    \end{algorithmic}  
\end{algorithm}

Additionally, this anomaly detection model (Model \Rmnum{2} in Table \ref{tab:optimModel}, or Model \ref{model:TBSD}) is theoretically compared with optimization models of RPCA and SSD as follows:

\vspace{-1em}
\begin{subequations}
\begin{flalign}
    & \text{RPCA method\citep{candes2011robust}:} \quad 
    \begin{aligned}
        & \underset{W}{\operatorname{argmin}} \quad \|W\|^2+\lambda \Vert L\Vert_* +\gamma\|S\|_1 \\
        & s.t. \quad Y = L + S + W
    \end{aligned}
    \label{model:RPCA} \\
    \cline{1-2}
    & \text{SSD method\citep{yan2017anomaly,yan2018real}:} \quad 
    \begin{aligned}
        & \underset{\theta, \theta_a}{\operatorname{argmin}} \quad \|e\|^2+\lambda \theta^T R\theta +\gamma\|\theta_a\|_1 \\
        & s.t. \quad Y = \widehat{C_{bg}} + \widehat{C_a} + e = B\theta + B_a\theta_a + e\\
    \end{aligned}
    \label{model:SSD} \\
    \cline{1-2}
    & \text{Proposed TBSD:} \quad 
    \begin{aligned}
        & \underset{\theta, \theta_t}{\operatorname{argmin}} \quad \|e\|^2+\lambda \theta^T R\theta +\gamma\|\theta_t\|_1 +\eta\Vert \widehat{C_a}\Vert_1 \\
        & s.t. \quad Y = \widehat{C_{bg}} + \widehat{C_{tex}} + \widehat{C_a} + e = B\theta + B_t\theta_t + \widehat{C_a} + e \\
    \end{aligned}
    \label{model:TBSD}
    \end{flalign}
\end{subequations}

where $\Vert\cdot\Vert^2, \Vert\cdot\Vert_1, \Vert\cdot\Vert_*$ are respectively defined as the l2-norm, l1-norm, and the nuclear-norm operators, which are separately denoted as $\Vert M\Vert^2=\sum_{ij}{M_{ij}^2}$, $\Vert M\Vert_1=\sum_{ij}{\vert M_{ij}\vert}$, and $\Vert M\Vert_*=\sum_i{\sigma_i(M)}$, with $\sigma_i(M)$ represents the $i^{th}$ singular value of $M$.

When compared with those benchmarks, the contributions of the proposed TBSD method on anomaly detection of textured images can be concluded as follows:
\begin{enumerate}
    \item This study primarily addresses the issue of misidentification between anomalies and normal textures in textured images, which is not theoretically highlighted and practically accomplished in other benchmarks.

    \item This study has provided a solid mathematical foundation with quasi-periodicity introduced so that image textures could be reasonably decomposed from original data. Accordingly, this study has also proposed a practical approach to learn textures and integrate them into optimization models for efficient anomaly detection.

    \item The proposed method can be easily implemented with stably satisfying anomaly detection performance in textured images, where other benchmarks may suffer from complicated parameter tuning work without guarantee on detection accuracy.
\end{enumerate}

\subsection{Computation Complexity Analysis}
The computation complexity of two operations are listed here for premises: 
\begin{enumerate}
    \item Inversion operation of a $n\times n$ matrix: $O(n^3)$
    \item Multiplication operation of a $m\times n$ matrix with another $n\times p$ matrix: $O(m\cdot n\cdot p)$
\end{enumerate}

Subsequently, here are some assumptions about data or parameters used in the method:
\begin{enumerate}
    \item Size of the defect-free image patch for training: $m\times n$.
    \item Size of the defect image for anomaly detection: $M\times N (m\le M, n\le N)$.
    \item Parameters to construct the smooth basis function $B$: $(k, L)$.
    \item Number, width, and rotation times of sampling lines in Algorithm \ref{algorithm:RotateSampling}: $(K_l, w_l, K_r)$.
    \item Number of learned texture basis functions: $K_t$.
\end{enumerate}
	
Based on all aforementioned premises and assumptions:

\textbf{Computation complexity to estimate components are:}
\begin{enumerate}
    \item Estimation of $C_{bg}$ in train/test images (Eq \ref{eq:smoothnessdecom2}): $O({max(m, n)}^3)/O({max(M, N)}^3)$
    \item Estimation of $C_{tex}$ based on sparse dictionary learning: $ O((K_t\cdot M\cdot N^2 ))$
\end{enumerate}

\textbf{Computation complexity of main algorithms are:}
\begin{enumerate}
    \item Low-rank Data Decomposition Process (Algorithm \ref{algorithm:LowRankDecomposite}): $O({max(m, n)}^3)$.
    \item LSERA Process (Algorithm \ref{algorithm:RotateSampling}): approximately $ O(K_r\cdot K_l\cdot w_l\cdot\sqrt{(m^2+n^2 )})$. 
    \item Anomaly Detection Process (Algorithm \ref{algorithm:DefectImageDecomposition}): $O({max(M, N)}^3)$.
\end{enumerate}
	
In summary, the computation complexity of the proposed TBSD method has a maximum accessibility of $O({max(m, n)}^3)$ in the training phase and $O({max(M, N)}^3)$ in the detection phase, which are highly correlated with the size of images chosen for corresponding processes. To compare with, the computation complexity of FFT is $O(N^2 logN)$, while the RPCA and SSD methods both have the computation complexity of $O({max(M, N)}^3)$ because of matrix operations. The slight increase in computation time would be acceptable considering the significant anomaly detection effects which are convinced by both simulated- and real-world experiments in the next two sections.



\section{Simulation Study} \label{s:simuexp} 
\begin{figure}[h]
    \centering
    \includegraphics[width=0.9\textwidth]{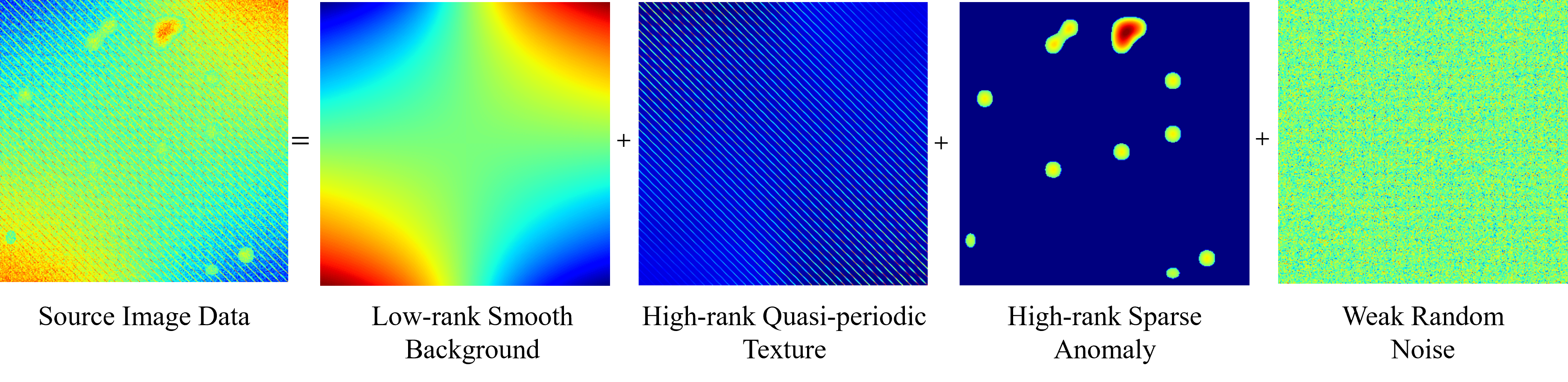}
    \caption{Schematic diagram of simulation image data components}
    \label{fig:simu_example}
\end{figure}

\
A simulation dataset is designed as Figure \ref{fig:simu_example} with a size of $344\times 351$ in pixel.

A defect-free simulation image, with cross-textures and a texture spacing of 10 pixels, is denoted as $Y_{cross[10]}^{simulate}$. This image is selected for TBFL process to learn the texture basis $B_t$. The quasi-periodicity detection based on Algorithm \ref{algorithm:QuasiPeriodicDetect}
is conducted on the texture estimator $\widehat{C_{tex}}$, which is obtained by Algorithm \ref{algorithm:LowRankDecomposite} on $Y_{cross[10]}^{simulate}$.
The results indicate the presence of two texture expansion directions including $45^\circ$ and $135^\circ$ in $Y_{cross[10]}^{simulate}$, aligning with real-world observations. Then $B_t$ is learned by Algorithm \ref{algorithm:RotateSampling} and \ref{algorithm:KNBN}.
Finally, the learned texture basis $B_t$ is input into the subsequent anomaly detection process (Algorithm \ref{algorithm:DefectImageDecomposition}) conducted on defect images with $(\lambda, \gamma, \eta, iterTimes) = (0.1, 0.2, 0.05, 1)$.

\begin{figure}[h]
    \centering
    \includegraphics[width=\textwidth]{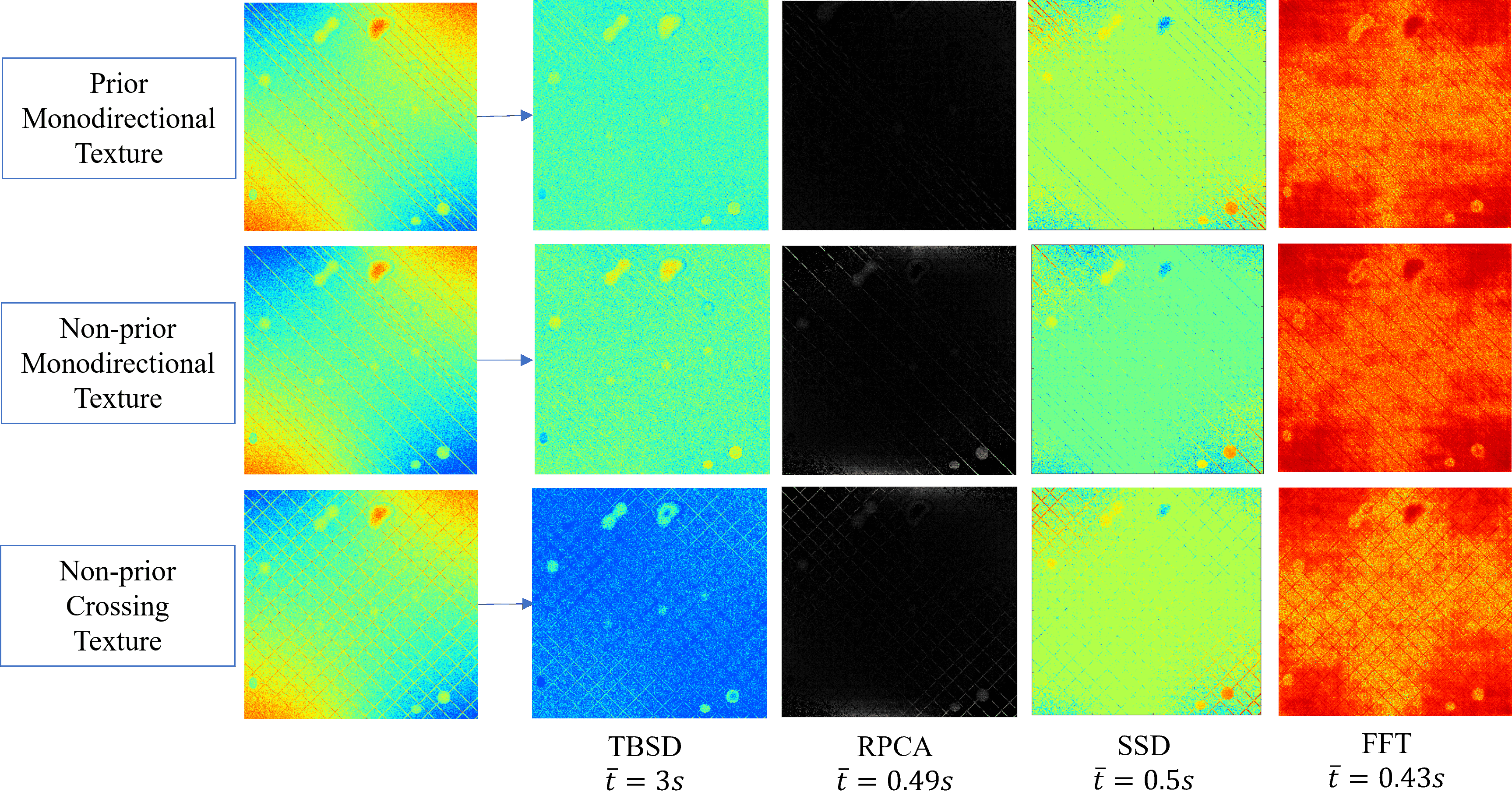}
    \caption{Schematic diagram of anomaly detection effect}
    \label{fig:simu_compareBenchmark}
\end{figure}

As results shown in Figure \ref{fig:simu_compareBenchmark}, the preset anomalies are separated with a relatively high accuracy, which means TBSD is practical for image anomaly detection. 
As for comparison, three benchmarks including RPCA, SSD, and FFT are selected to deal with the same images. 
The proposed TBSD method achieves the best detection results with the most complete anomalies extracted and misidentified textures reduced to a large extent. Other methods all possess varying degrees of misidentification and relatively poor conformity with the underlying truth. In summary, the proposed TBSD method provides an efficient approach to enhance accuracy without significantly increasing time on textured image anomaly detection.


For quantitative comparison, two metrics including the true positive rate (TPR) and the false positive rate (FPR) are introduced:
\begin{equation}
    TPR = \frac{TP}{TP+FN}, \qquad
    FPR = \frac{FP}{FP+TN},
    \label{eq:performance_metrics}
\end{equation}
where TP is the number of defect pixels correctly detected; FN is the number of defect pixels that are not detected; FP is the number of normal pixels that are incorrectly detected as anomalies; TN is the number of normal pixels correctly detected. 

Additionally, Algorithm \ref{algorithm:ClosedCurveForm} is applied onto anomaly detection results in Figure \ref{fig:simu_compareBenchmark}. This algorithm can utilize detected defect pixels to generate closed defect areas, which are quantitatively compared with the underlying truth. 
The detailed results are listed in Table \ref{table:simuNumCompare}, which provides a quantitative support for the graphical results. It is shown that TBSD method achieves the best anomaly detection performance among all three types of defect simulation textured images when compared with benchmarks.

\begin{table}[H]
    \centering
    \caption{Simulation numerical experiment results of anomaly detection methods}
    \label{table:simuNumCompare}
    \begin{tabular}{c|c|c|c|c|c|c|c|c}
      \hline
      \multirow{2}{*}{\thead{\large Texture \\ \large Pattern}} & \multicolumn{2}{c}{TBSD} \vline & \multicolumn{2}{c}{RPCA} \vline & \multicolumn{2}{c}{SSD} \vline & \multicolumn{2}{c}{FFT} \\
      \cline{2-9}
       & TPR & FPR & TPR & FPR & TPR & FPR & TPR & FPR \\
      \hline
      \thead{\large Prior One-direction} & \textbf{0.412} & {0.095} & {0.297} & 0.248 & 0.268 & \textbf{0.072} & 0.079 & 0.101\\
      \hline
      \thead{\large Non-prior One-direction} & \textbf{0.547} & \textbf{0.079} & 0.283 & 0.249 & 0.337 & {0.084} & 0.063 & 0.101 \\
      \hline
      \thead{\large Non-prior Crossing} & \textbf{0.331} & {0.096} & 0.260 & 0.250 & 0.268 & \textbf{0.071} & 0.057 & 0.102 \\
      \hline
      Avg. & \textbf{0.430} & {0.090} & 0.280 & 0.249 & 0.291 & \textbf{0.075} & 0.066 & 0.101 \\
      \hline
    \end{tabular}
\end{table}

\section{Case Study} \label{s:realexp} \
Real-world experiments are conducted on industrial images from the ``wood" sub-dataset in a publicly available MVTec AD dataset\citep{bergmann2021mvtec}. 
With the quasi-periodicity detection result indicating that the textures expand horizontally, the TBFL process (Table \ref{tab:optimModel} \Rmnum{1}; Algorithm \ref{algorithm:LowRankDecomposite}, \ref{algorithm:RotateSampling}) is applied on a selected image patch to learn texture basis functions $B_t$. This patch is segmented from the \textit{075} defect-free image $Y_{075}^{good}$ in the ``good" category of the ``wood" sub-dataset. It is obtained by dividing $Y_{075}^{good}$ into 16 equal parts and separating the (0, 0) part in the first row and the first column. The entire patch is denoted as $Y_{075, [0, 0]}^{good}$.

In like manner, following notations like $Y_{003}^{hole}$ and $Y_{000, [1, 1]}^{hole}$ respectively represent the \textit{003} defect image in the ``hole" category of the ``wood" sub-dataset, and the (1, 1) patch in the second row and the second column of equally 16-segmented $Y_{000}^{hole}$.

\subsection{Preliminary Experiment}

\begin{figure}[h]
    \centering
    \includegraphics[width=0.8\linewidth]{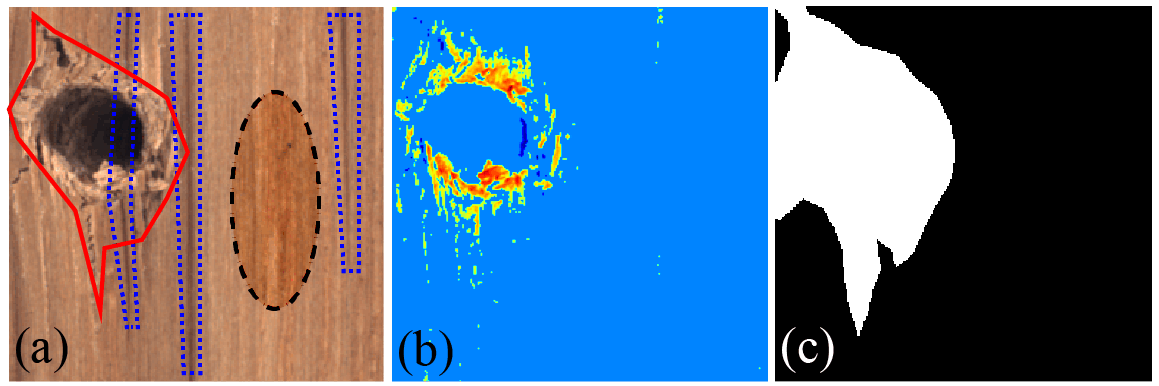}
    \caption{\textcolor{black}{Example of proposed TBSD method for anomaly detection on real-world image with large-size anomalies: (a) Original textured image, where the \textcolor{red}{solid line} marks the anomalies, \textcolor{blue}{dotted line} marks the textures, \textbf{circular mask} marks part of the background; (b) Anomalies detected by the proposed TBSD method; (c) Underlying truth of anomalies}}
    \label{fig:TBSD_anomaly_detection}
\end{figure}

A defect image patch $Y_{000, [1, 1]}^{hole}$ is selected for the anomaly detection process (Table \ref{tab:optimModel} \Rmnum{2}; Algorithm \ref{algorithm:DefectImageDecomposition}) with $(\lambda, \gamma, \eta, iterTimes) = (0.1, 0.2, 0.05, 1)$. 
An excellent performance is accomplished in Figure \ref{fig:TBSD_anomaly_detection}(b) with both clear distinction from normal textures and accurately separated anomalies. 
This preliminarily proves the outstanding ability of the proposed TBSD method, which could conduct efficient anomaly detection with texture misidentification prevented at the same time, and high tolerance on the scale of anomalies which violates preset sparsity.

\begin{figure}[h]
    \centering
    \includegraphics[width=\textwidth]{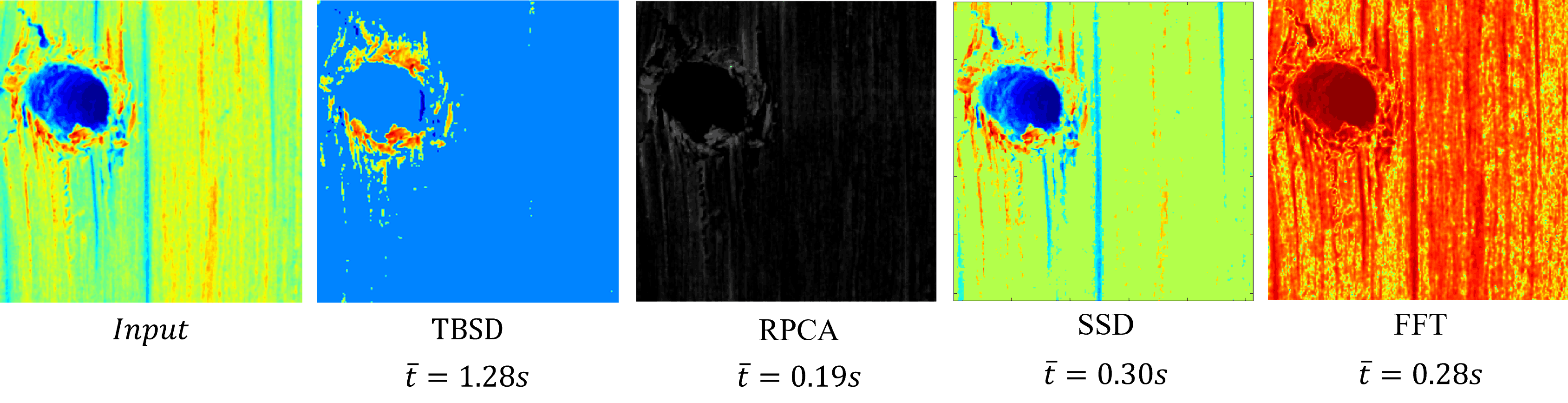}
    \caption{Comparison of anomaly detection effect based on $Y_{000, [1, 1]}^{hole}$}
    \label{fig:real_compareBenchmark}
\end{figure}

Additionally, performance of benchmarks are compared in Figure \ref{fig:real_compareBenchmark}. The experiment conducted on the real-world dataset yields similar results to the simulation study (Section \ref{s:simuexp}) with respect to anomaly detection performance of each method, which also conform to their theoretical limitations discussed in Section \ref{subsec:ADprocess}.
\begin{enumerate}
    \item The proposed TBSD demonstrates successful anomaly detection with high consistency with underlying truth. It also reduces misidentified normal textures to a great extent with a slightly longer but acceptable detection time.
    \item The RPCA and FFT methods possess fast speeds but poor anomaly detection effect. RPCA captures too few anomalies, while FFT extracts excessive textures and background due to non-optimal filtering. 
    \item The SSD method
    captures relatively complete anomalies, while it also misidentifies more normal textures as anomalies. 
\end{enumerate}


\subsection{Expanded Experiment}

Expanded numerical experiments are conducted on three randomly selected images 
including $Y_{000}^{hole}$, $Y_{003}^{hole}$, and $Y_{009}^{hole}$. Each of those images is quartered both horizontally and vertically into 16 sub-images and separately processed using the TBSD method.
Similar with the simulation study, anomaly detection results are also processed by Algorithm \ref{algorithm:ClosedCurveForm} to support quantitative comparison with respect to $TPR$ and $FPR$ metrics (Eq. \ref{eq:performance_metrics}). 
The same parameter $(\phi_{B/T}, \phi_A)=(0.5, 2\%)$ is applied to ensure the comparison to reflect more about TBSD's superiority rather than impact of parameter tuning. However, the parameter selection also plays a crucial role in TBSD's performance though it does possess relatively strong robustness to non-optimal parameter settings.

\begin{figure}[h]
    \centering
    \includegraphics[width=.75\textwidth]{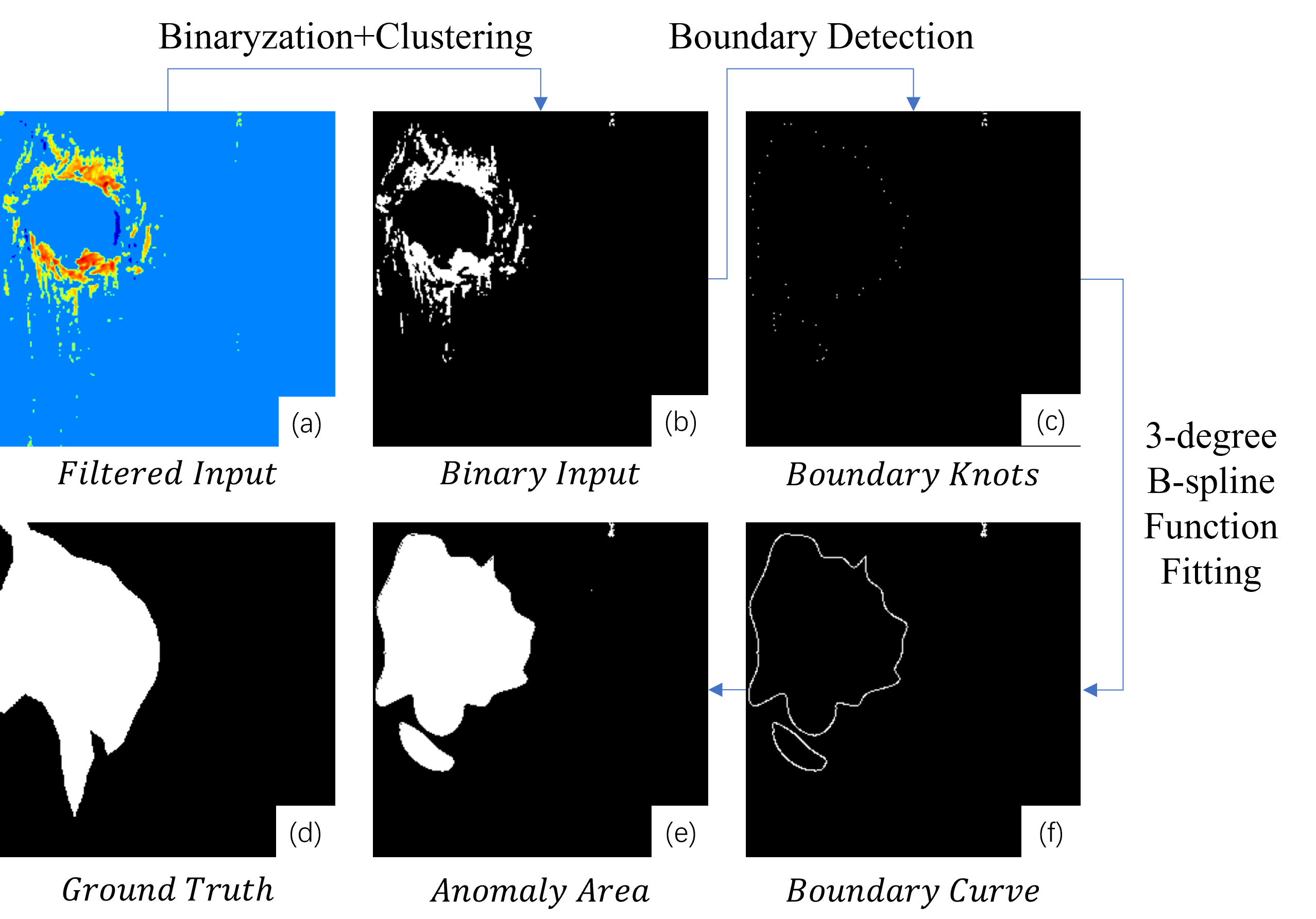}
    \caption{Visualization of anomaly detection effect based on $Y_{000, [1, 1]}^{hole}$}
    \label{fig:real_closeBoundary}
\end{figure}

Take the anomaly detection result of $Y_{000, [1, 1]}^{hole}$ as an example (Figure \ref{fig:real_closeBoundary}).
The proportion of detected anomalies is evaluated to about 6\%, which far exceeds a preset threshold $\phi_A=2\%$ and indicates the defect issue within that image. 
Closed defect areas in Figure \ref{fig:real_closeBoundary}(e) is quantitatively measured with $TPR\approx 78.44\%$ and $FPR\approx 2.16\%$, which demonstrate that almost 80\% of anomalies are detected with misidentification less than 3\%. 

\begin{figure}[H]
    \centering
    \includegraphics[width=.8\textwidth]{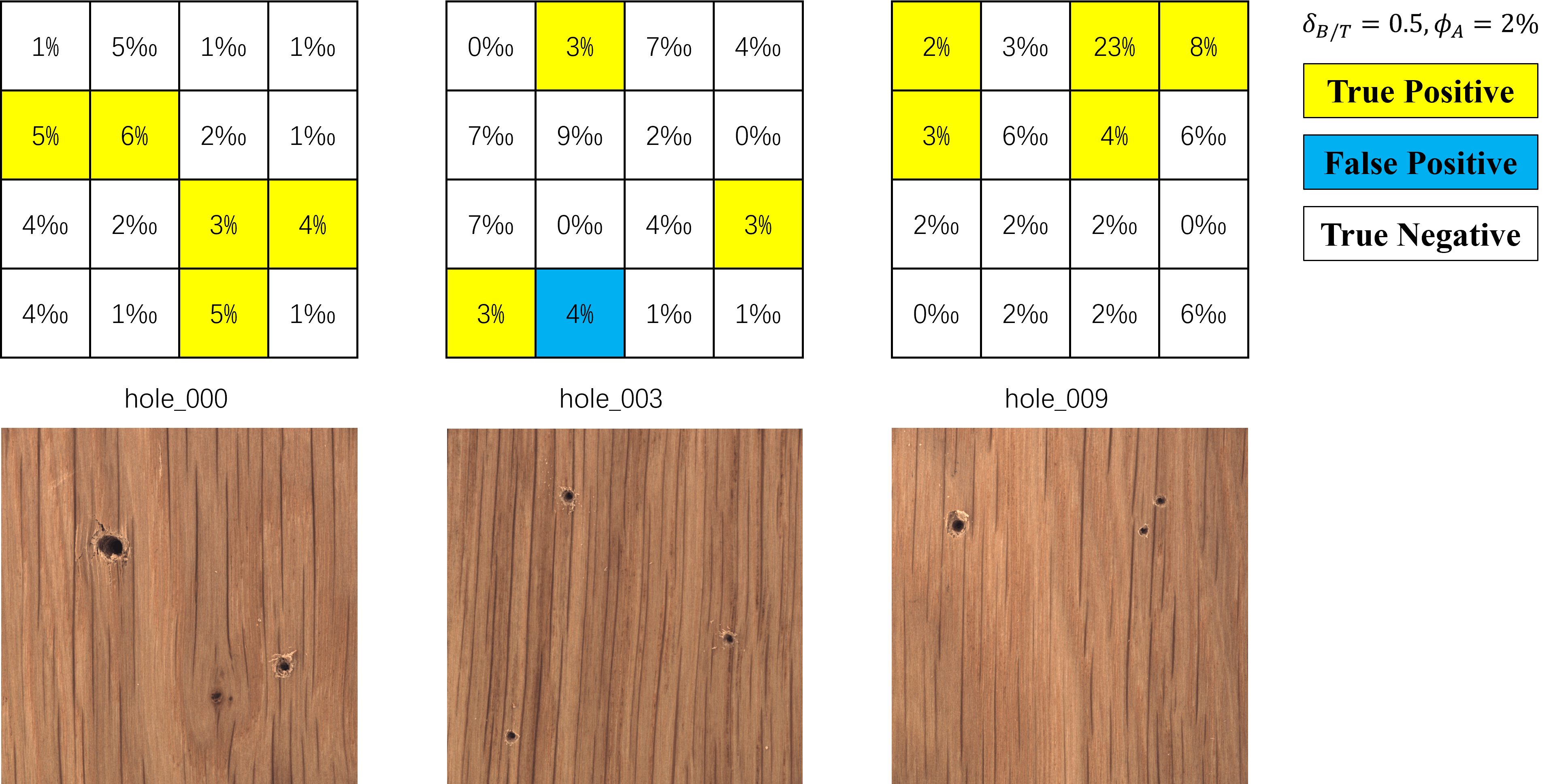}
    \caption{Positive proportions of patches' anomaly detection results in $Y_{000}^{hole}$, $Y_{003}^{hole}$, $Y_{009}^{hole}$
    (Top: positive pixels' proportion; Bottom: corresponding real-world images)}
    \label{fig:real_ADresultScale}
\end{figure}

Furthermore, Table \ref{tab:real_numExpResult} summarizes the quantitative experiment results in terms of $TPR$ and $FPR$. Figure \ref{fig:real_ADresultScale} has illustrated the proportions of detected defect pixels in $Y_{000}^{hole}$, $Y_{003}^{hole}$, and $Y_{009}^{hole}$ after applying TBSD method. 

\begin{table}[H]
    \centering
    \setlength{\tabcolsep}{3pt}
    \caption{Numerical experimental statistics of defect patches in $Y_{000}^{hole}$, $Y_{003}^{hole}$, $Y_{009}^{hole}$}
    \label{tab:real_numExpResult}
    \begin{tabular}{c|c|ccc|cc|cc|cc}
    \toprule
    \multirow{3}{*}{\thead{Image\\ Id \\ \:}} & \multirow{3}{*}{\thead{Patch \\ Index\\ \:}} & \multicolumn{3}{c}{TBSD} \vline & \multicolumn{2}{c}{RPCA} \vline & \multicolumn{2}{c}{SSD} \vline & \multicolumn{2}{c}{FFT} \\
    \cline{3-11}
    & & {\small $(D_{max}, \sigma)$} & TPR & FPR & TPR & FPR & TPR & FPR & TPR & FPR \\
    \midrule
    \multirow{5}{*}{000} & (1, 0) & (40, 20) & \textbf{0.649} & \textbf{0.045} & {0.430} & 0.243 & 0.319 & 0.248 & 0.242 & {0.095} \\
    \cline{2-11}
    & (1, 1) & (80, 5) & \textbf{0.784} & \textbf{0.022} & 0.342 & 0.221 & {0.357} & 0.217 & 0.324 & {0.030} \\
    \cline{2-11}
    & (2, 2) & (20, 20) & 0.244 & \textbf{0.005} & 0.658 & 0.247 & \textbf{0.917} & 0.244 & {0.879} & {0.091} \\
    \cline{2-11}
    & (3, 2) & (30, 25) & 0.323 & \textbf{0.013} & 0.225 & 0.251 & {0.749} & 0.239 & \textbf{0.776} & {0.085} \\
    \cline{2-11}
    & (2, 3) & (30, 5) & \textbf{0.770} & \textbf{0.011} & 0.250 & 0.250 & 0.412 & 0.234 & {0.459} & {0.064} \\
    \midrule
    \multirow{3}{*}{003} & (3, 0) & (20, 5) & \textbf{0.667} & \textbf{0.047} & 0.314 & 0.248 & 0.343 & 0.247 & {0.402} & {0.090} \\
    \cline{2-11}
    & (0, 1) & (40, 5) & \textbf{0.503} & \textbf{0.054} & 0.260 & 0.249 & {0.323} & 0.242 & 0.293 & {0.080} \\
    \cline{2-11}
    & (2, 3) & (40, 5) & \textbf{0.349} & \textbf{0.046} & 0.249 & 0.250 & {0.329} & 0.244 & 0.270 & {0.087} \\
    \midrule
    \multirow{5}{*}{009} & (0, 0) & (40, 25) & \textbf{0.658} & \textbf{0.001} & 0.404 & 0.247 & 0.391 & 0.247 & {0.541} & {0.090} \\
    \cline{2-11}
    & (1, 0) & (80, 5) & \textbf{0.896} & \textbf{0.015} & 0.321 & 0.244 & 0.389 & 0.240 & {0.576} & {0.067} \\
    \cline{2-11}
    & (0, 2) & (20, 25) & \textbf{0.362} & \textbf{0.000} & 0.218 & 0.250 & 0.350 & 0.249 & {0.358} & {0.097} \\
    \cline{2-11}
    & (1, 2) & (20, 20) & \textbf{0.650} & \textbf{0.002} & 0.248 & 0.250 & 0.344 & 0.248 & {0.440} & {0.093} \\
    \cline{2-11}
    & (0, 3) & (20, 20) & \textbf{0.685} & \textbf{0.024} & 0.377 & 0.247 & 0.358 & 0.248 & {0.392} & {0.094} \\
    \midrule
    \multicolumn{3}{c}{Performance Avg.} \vline & \textbf{0.580} & \textbf{0.022} & 0.330 & 0.246 & 0.429 & 0.242 & {0.458} & {0.082} \\
    \midrule
    \multicolumn{3}{c}{Performance Std.} \vline & 0.196 & 0.019 & \textbf{0.116} & {0.008} & {0.177} & \textbf{0.008} & 0.185 & 0.018 \\
    \midrule
    \multicolumn{3}{c}{\thead{Five Best \\ Performance}} \vline & \thead{\textbf{0.761}\\$\pm$ \textbf{0.082}} & \thead{\textbf{0.024}\\$\pm$ 0.010} & \thead{0.442\\$\pm$ {0.112}} & \thead{0.241\\$\pm$ \textbf{0}} & \thead{0.572\\$\pm$ 0.220} & \thead{0.241\\$\pm$ {0.001}} & \thead{{0.646}\\$\pm$ 0.156} & \thead{{0.080}\\$\pm$ 0.002} \\
    \midrule
    \multicolumn{3}{c}{\thead{Five Worst \\Performance}} \vline & \thead{\textbf{0.356}\\$\pm$ 0.084} & \thead{\textbf{0.024}\\$\pm$ \textbf{0.004}} & \thead{0.238\\$\pm$ {0.014}} & \thead{0.250\\$\pm$ 0.010} & \thead{{0.332}\\$\pm$ \textbf{0.010}} & \thead{0.246\\$\pm$ {0.009}} & \thead{0.297\\$\pm$ 0.041} & \thead{{0.078}\\$\pm$ 0.019} \\
    \bottomrule
    \end{tabular}
\end{table}

Except for one misjudgement in $Y_{003}^{hole}$, all defect patches are accurately detected with relatively significant differences from other defect-free patches.
The proposed TBSD method exceeds other benchmarks on both the average level and top-performing level across randomly selected defect images. 
All the aforementioned conclusions strongly corroborate the superior effectiveness and reliability of TBSD method for image anomaly detection.

\begin{figure}[h]
    \centering
    \includegraphics[width=\textwidth]{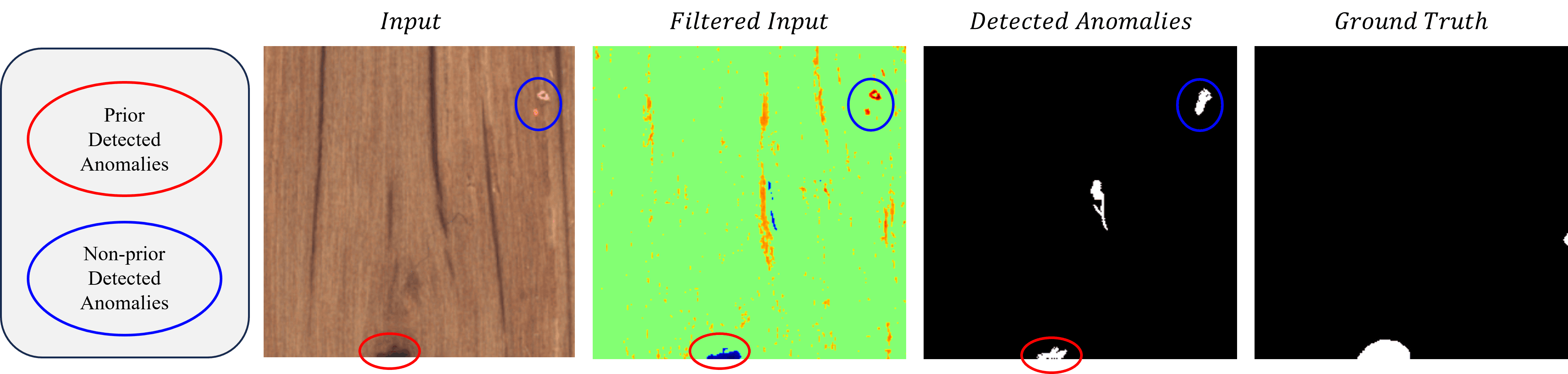}
    \caption{TBSD's ability to detect targeted defects and non-prior anomaly types}
    \label{5_AnomalyPredict}
\end{figure}

It is also worth noting that some of the "misidentified defect pixels" may represent potential anomalies that had not been labeled. This assertion finds some confirmation through the comparison of anomaly detection effects as presented in Figure \ref{5_AnomalyPredict}. 
It is implied that the proposed TBSD method exhibits a certain predictive capability for noble anomaly types where prior information is lacking.

As supplementary, it is important to note that the universal parameter set $(\phi_{B/T}, \phi_A) = (0.5, 2\%)$ 
may not necessarily be the optimal parameter combination for anomaly detection in each image patch. 
As illustrated in Figure \ref{5_FiveBest} and \ref{5_FiveWorst}, an improper choice of parameters can led to partially residual textures in detected anomalies $C_a$
, which is more pronounced in Figure \ref{5_FiveWorst}. 
This suggests that significant scope exists for adjusting parameters even between different image patches captured from the same source image.
With appropriate cost, identifying an optimal parameter combinations would further improve the anomaly detection results of the proposed TBSD method on textured images.

\section{Conclusion} \label{s:conc}
In the proposed TBSD method, efficient anomaly detection is conducted on textured images with smooth background and sparse anomalies. Problem of misidentifying textures as anomalies is prevented both theoretically and practically when compared with benchmarks. With quasi-periodicity introduced into textures, the method firstly learns a set of texture basis functions using a small number of defect-free images. Then the subsequent anomaly detection process utilizes the learned texture basis for texture reconstruction, which leads to improved detection performance on defect images. The method's accuracy and superiority are sufficiently evaluated through experimental verification using both pre-constructed simulation dataset and real-world industrial dataset, which highlights its feasibility, effectiveness and efficiency.

As far as we are concerned, an appropriate parameter choice can evidently improve the method's performance though it does possess robustness to various parameters.
Optimal parameters may vary across different image patches depending on the circumstances. 
There still needs further research to properly deal with the sensitivity of the TBSD method to relevant parameters.



\section*{Data and code Availability}
The code and data are available via the link https://github.com/LaMiracle/TBSD.  

\section*{Acknowledgements}
The work is partially supported by the Beijing Natural Science Foundation (No. 3244032, L241039), and partially supported by the National Natural Science Foundation of China (No. 92467302, U24B20116) as well as supported by the Opening Project Fund of Materials Service Safety Assessment Facilities. 
The authors thank the editors and reviewers for helping improve the paper significantly.

\bibliographystyle{apalike}
\bibliography{references}
\newpage


\begin{appendix}
\input{appendix/Nomenclatures}

\input{appendix/algorithm}
\input{appendix/prove}

\input{appendix/figures}
\end{appendix}

\end{document}

%% file: appendix/Nomenclatures.tex
\section{Nomenclatures}
\subsection{Nomenclatures with respect to Quasi-periodicity}
\begin{centering}
\setlength\LTleft{0pt}
\setlength\LTright{0pt}
\begin{longtable}{llm{8.25cm}}
    \caption{Nomenclatures w.r.t. Quasi-periodicity} \\
    \hline \textbf{Notation} & \textbf{Definition} & \textbf{Description} \\ \hline
    \endfirsthead

    \multicolumn{3}{c}%
    {{\bfseries \tablename\ \thetable{} -- continued from previous page}} \\
    \hline \textbf{Notation} & \textbf{Definition} & \textbf{Description} \\ \hline
    \endhead

    \hline \multicolumn{3}{l}{{Continued on next page}} \\
    \endfoot

    \bottomrule
    \endlastfoot
    
    $s$ & $s\in \mathbb{R}$ & Real-valued color intensity sampled from targeted image data;\\
    
    $S$ & $\{s_1, s_2, s_3, \dots, s_n\}\in \mathbb{R}^1$ & One-dimensional (1D) signal set; \\
    
    $S^2$ & $\{S_1, S_2, S_3, \dots, S_n\}\in \mathbb{R}^2$ & Two-dimensional (2D) signal set;
    \\

    $S^{(k)}$ & * & A marker of the $k^{th}$ component which makes up a composite 1D signal set; \\

    $S^{[k]}$ & * & A marker of the $k^{th}$ composite 1D signal set which generates a 2D signal set; \\

    $S^{\widetilde{t_m}}$ & $\{S_{t_m}, S_{t_m+1}, \dots, S_{t_{m+1}}\}$ & A marker of the $m^{th}$ segment from a 1D signal set; \\

    $T$ & $T\in \mathbb{Z}^+$ & Periodic length of a 1D periodic signal set; \\

    $T^2$ & $T^2\in {\mathbb{Z}^+}^2$ & Periodic size of a 2D periodic signal set; \\

    $S_T$ &
    $
    \left\{\begin{aligned}
        &s_1, s_2, s_3, \dots, s_T \\
        &\vert s_j \in S, 1<T<n
    \end{aligned}\right\}
    $
    & Periodic mode of a 1D periodic signal set $S$; \\
    \addlinespace[1em]

    $S_{\tilde{T}}$ & 
        $
        \left\{
        \begin{aligned}
            & s_{a+1}, s_{a+2}, \dots, s_{a+T} \\
            & \left| 
            \begin{aligned}
                & s_j \in S, \, 1 < T < n, \\
                & 0 \leq a \leq n - T
            \end{aligned}
            \right.
        \end{aligned}
        \right\}
        $
    & 
    Quasi-periodic mode of a 1D quasi-periodic signal set $S$; \\

    $S_{\{t_M\}}$ 
    & 
        \small{$\left\{\begin{aligned}
            & S^{\widetilde{t_m}}, S^{\widetilde{t_{m+1}}}, \dots, S^{\widetilde{t_{M-1}}} \\ 
            &\vert \sum_{m=1}^M{t_m} = n, t_m \in \mathbb{Z}^+
            \end{aligned}\right\}
        $}
    & A segmentation which divides the 1D signal set $S$ into $M$ parts of $t_m$ length subsequently; \\

    $P^*$ & * & An operator which obtains the least common multiple of elements in the targeted set $P$;\\

    $[S]_{:T}$ 
    & 
        $\left\{\begin{aligned}
            \{& s_1, s_2, s_3, \dots, s_T \\
            & \vert s_j \in S\}
        \end{aligned}\right\}$
    & An operator which intercepts the first $T$ elements in the targeted 1D set $S$; \\

    $[S^2]_{(:T_1, :T_2)}$ & 
    \(
    \begin{array}[t]{@{}l@{}}
         [\{[S_j]_{:T^{[1]}}\}]_{:T^{[2]}}
    \end{array}
    \)
    & An operator which intercepts the first $T_1 \times T_2$ submatrix in the targeted 2D set $S^2$; \\

    $do(\cdot)$ & $S=do(S_T, \mathbb{F}_n)$ & An operator which creates a 1D periodic signal set $S$ with essential elements including $T$, $S_T$, and the corresponding definition domain $\mathbb{F}_n$; \\

    $quasi(\cdot)$ & $S=quasi(S_{\tilde{T}}, S_{\{t_M\}}, \sigma, \mathbb{F}_n)$ & An operator which creates a 1D quasi-periodic signal set $S$ with essential elements including $T$, $S_{\tilde{T}}$, $S_{\{t_M\}}$, $\sigma$, and $\mathbb{F}_n$; \\

    $\mathbb{R}$ & * & The real number field; \\
    $\mathbb{Z}$ & * & The positively integral number field; \\
    $\mathbb{F}_n$ & $\{i\in \mathbb{Z}^+ \vert 1\le i\le n\}$ & The definition domain of any signal sets, where $n$ represents the total number and $i$ represents the index of signals existing in the set.
    \label{tab:s3notations}
\end{longtable}
\end{centering}

\newpage

\subsection{Nomenclatures with respect to Optimization Model}
\begin{center}
\begin{longtable}{llm{9cm}}
    \caption{Nomenclatures w.r.t. Optimization Model} \\
    \hline \textbf{Notation} & \textbf{Type} & \textbf{Description} \\ \hline
    \endfirsthead

    \multicolumn{3}{c}%
    {{\bfseries \tablename\ \thetable{} -- continued from previous page}} \\
    \hline \textbf{Notation} & \textbf{Type} & \textbf{Description} \\ \hline
    \endhead

    \hline \multicolumn{3}{l}{{Continued on next page}} \\
    \endfoot

    \bottomrule
    \endlastfoot

    $Y$ & Image data & Denotation of the original image data serving as workspace for the proposed data decomposition method \\
    $C_{bg}$ & Image component & Denotation of the low-rank smooth background within images \\
    $C_{tex}$ & Image component & Denotation of the high-rank quasi-periodic textures within images \\
    $C_{a}$ & Image component & Denotation of the high-rank sparse anomalies within images \\
    $e$ & Image component & Denotation of the random noise except the other components within images \\

    $C_{tex}^{(d)}$ & Texture feature & Sub-texture along a specific texture expansion direction $d$ after dimensional-reduction \\
    $D$ & Vector set & Set of vector directions where sub-texture expands and possesses quasi-periodicity \\

    $B_t^{(d)}$ & Basis function & Learned basis function for the quasi-periodic sub-texture along a specific texture expansion direction $d$. The learning process is driven by Algorithm \ref{algorithm:KNBN} \\
    
    $B$ & Basis function & Pre-constructed basis functions for the smooth background component within images, a 3-degree B-spline curve for example \\
    $B_t=\{B_t^{(d)}\vert d\in D\}$ & Basis function & Learned basis functions for the quasi-periodic texture component within images, an orthogonal matrix learned from limited defect-free images \\

    $\theta$ & Coefficient matrix & Matrix multiplier for generating the smooth background component with corresponding pre-constructed basis functions \\
    $\theta_t=\{\theta_t^{(d)}\vert d\in D\}$ & Coefficient matrix & Matrix multiplier for generating the quasi-periodic component with corresponding pre-constructed basis functions \\
    $\hat{\theta}$ & Estimator & Estimator obtained by iterative algorithms for the smooth-basis coefficient matrix multiplier \\
    $\hat{\theta_t}=\{\widehat{\theta_t^{(d)}}\vert d\in D\}$ & Estimator & Estimator obtained by iterative algorithms for the texture-basis coefficient matrix multiplier \\

    $\widehat{C_{bg}} = B\hat{\theta}$ & Estimator & Low-rank smooth background's estimator which is obtained by the proposed TBSD method driven by iterative data decomposition \\
    $\widehat{C_{tex}} = B_t \hat{\theta_t}$ & Estimator & High-rank quasi-periodic textures' estimator which is obtained by the proposed TBSD method driven by iterative data decomposition \\
    $\hat{C_{a}}$ & Estimator & High-rank sparse anomalies' estimator which is obtained by conducting subtraction on the original image with the estimated background and textures \\

    $\|\cdot\|$ & Operator & L2 norm operator \\
    $\|\cdot\|_1$ & Operator & L1 norm operator \\

    $\lambda$ & Weight parameter & $0<\lambda\le 1$ is recommended for background smoothness controlling \\
    $\gamma$ & Weight parameter & $0<\gamma\le 1$ is recommended for texture sparsity controlling \\
    $R$ & Roughness matrix & Estimated by $R=D^TD$ with $D$ expressed in Equation \ref{eq:D}\\

    $\lambda \theta^T R\theta$ & Penalty & A penalty term on difference among adjacent pixels to constrain on the smoothness of image background\footnote{Here follows the idea of Yan et al.\cite{yan2017anomaly} to set a penalty term for the smooth background component.} \\
    $\gamma\|\theta_t\|_1$ & Penalty & A penalty term on the sparsity of the solved coefficient matrix for texture estimation, which corresponds to sparsely distributed textures \\

    \label{tab:s4notations}
\end{longtable}
\end{center}

\vspace{-4em}
\begin{equation}
    D=
    {\begin{bmatrix}
    1 & -1 &  &  \\
      & \ddots & \ddots &  \\
      &  & 1 & -1 \\
    \end{bmatrix}}_{(k+L-1)\times(k+L-1)},
    \label{eq:D}
\end{equation}
where $D$ represents the difference matrix of $I_{k+L-1}$\footnote{$I_m$ denotes an identify matrix with size of $m\times m$ }, with $L=3$ signifying the utilization of a 3-degree B-spline curve and $k$ reflecting the smoothness of the background curve.


%% file: appendix/algorithm.tex
\section{Other Available Algorithms}
\label{a:algo}
\begin{algorithm}[H]
    \caption{Linear Sampling in Equal Rotation Angle}
    \label{algorithm:RotateSampling}
    \begin{algorithmic}[1]
    \Require Source image: $Y$, Image dimensions: $m \times n$, Sampling width: $l_w$, Gap between centers: $l_g$, Max rotations: $maxRotate$
    \Ensure Sampling results: $allSamples$
    \Function{RotateSampling}{$Y, m, n, l_w, l_g, maxRotate$}
        \State $P_{datum}^0, iRotate, allSamples \gets (x_d^0, y_d^0), 0, \emptyset$
        \While{$iRotate < maxRotate$}
            \State $\alpha \gets \frac{iRotate}{maxRotate}\cdot \pi$
            \State $maxStep, samples, iGap \gets \max\left(\left|\frac{m}{\sin(\alpha)}\right|, \left|\frac{n}{\cos(\alpha)}\right|\right), \emptyset, 0$
            
            \While{sampling within boundaries of $Y$}
                \State $sampleUnit, iStep \gets \emptyset, 0$
                \For{ texture expansion direction $d^p \in \{\alpha + \frac{\pi}{2}, \alpha + \frac{3\pi}{2}\}$}
                \State $x_{d}^{iGap}, y_{d}^{iGap} \gets x_d^0 - d^p \cdot iGap \cdot l_g \cdot \cos(\alpha), y_d^0 - d^p \cdot iGap \cdot l_g \cdot \sin(\alpha)$
                \State $P_{datum}^{iGap} \gets (x_{d}^{iGap}, y_{d}^{iGap})$
                \For{ texture extension direction $d^t \in \{\alpha, \alpha + \pi\}$}
                        \State $P_{now}^{iGap, iStep}(x) \gets x_d^{iGap} - d^t \cdot iStep \cdot \sin(\alpha)$
                        \State $P_{now}^{iGap, iStep}(y) \gets y_d^{iGap} - d^t \cdot iStep \cdot \cos(\alpha)$
                        \State $P_{now}^{iGap, iStep} \gets (P_{now}^{iGap, iStep}(x), P_{now}^{iGap, iStep}(y))$
                        \State $sampleUnit \gets$ \{points within $l_w$ of $P_{now}^{iGap, iStep}$ along $d^p$\}
                        \State $samples \gets samples \bigcup sampleUnit$
                \EndFor
                \State $iStep \gets iStep +1$
                \State $iGap \gets iGap + 1$
                \EndFor
            \EndWhile
        
            \State $iRotate, allSamples \gets iRotate + 1, allSamples \bigcup samples$
        \EndWhile
        \Return $allSamples$
        \EndFunction
    \end{algorithmic}
\end{algorithm}
\begin{algorithm}[H]
    \caption{Quasi-periodicity Detection based on 1D Discrete Signal Set}
    \label{algorithm:QuasiPeriodicDetect}
    \begin{algorithmic}[1]
        \Require Unit rotation angle: $\theta$, 1D discrete signal set under each rotation angle: $[S_{k \theta} \vert 0\le k\le (\pi/\theta - 1)]$, Threshold ratio: $q$
        \Ensure Texture extension directions set: $D^e$, Texture expansion directions set: $D^p$, Standard deviation detection threshold: $\phi$
        \Function {QuasiPeriodicDetect}{$\theta, [S_{k \theta} \vert 0\le k\le (\pi/\theta - 1)], q$}
        \State $D^e \gets \emptyset, D^p \gets \emptyset$
        \State $S^{quasi} \gets \emptyset$
        \State $maxRotate \gets \pi/\theta$
        \State $iRotate \gets 0$
        \While {$iRotate < maxRotate$}
        \State $\alpha_1 \gets iRotate \cdot \theta$
        \If{$iRotate \le maxRotate / 2$}
        \State $\alpha_2 \gets \alpha_1 + \pi/2$
        \Else \State $\alpha_2 \gets \alpha_1 - \pi/2$
        \EndIf
        \State Sampling $S_{\alpha_1}, S_{\alpha_2}$ along directions $\alpha_1, \alpha_2$
        \State $S^{quasi} \gets S^{quasi} \bigcup \{[\alpha_1, \alpha_2, std(S_{\alpha_1}) - std(S_{\alpha_2})]\}$
        \EndWhile
        \State $\phi \gets q\cdot (max(abs(S^{quasi})) - min(abs(S^{quasi}))) + min(abs(S^{quasi}))$
        \For {$\alpha_1, \alpha_2, dev$ \ in \ $S^{quasi}$}
        \If {$dev > \phi$}
        \State $D^e \gets D^e \bigcup \{\alpha_1\}$
        \State $D^p \gets D^p \bigcup \{\alpha_2\}$
        \EndIf
        \EndFor
        \State \Return $D^e, D^p, \phi$
        \EndFunction
    \end{algorithmic}
\end{algorithm}
\begin{algorithm}[H]
    \caption{Texture Basis Function Learning based on KNBN Clustering}
    \label{algorithm:KNBN}
    \begin{algorithmic}[1]
        \Require Image texture estimator: $\widehat{C_{tex}}$, Texture expansion directions set $D^p$, Maximum threshold of basis function capacity: $K$, 
        Width of pixel neighborhood: $l$
        \Ensure Texture basis function set: $B_t$
        \Function {KNBN}{$\widehat{C_{tex}}, K, l$}
        \State $B_t, tex \gets \emptyset, \emptyset$
        \For {$d\in D^p$}
        \State $B_t^{(d)} \gets \emptyset$
        \While {$\widehat{C_{tex}} \neq \emptyset$}
        
        \State $avaiSeeds, tex \gets$ element[0] \ in \ $\widehat{C_{tex}}, \emptyset$
        \While {$avaiSeeds \neq \emptyset$}
        \State $seed \gets$ element[0] \ in \ $avaiSeeds$
        \State $neighbors \gets\{$bordered \ points \ within \ $l$ \ along \ $d$ \ from \ $seed\}$
        \If {$neighbors \neq \emptyset$}
        \State $tex, avaiSeeds \gets tex \bigcup seed, avaiSeeds \setminus seed$
        \State $avaiSeeds \gets avaiSeeds \bigcup neighbors$
        \If {$len(tex) \ge K$}
        \State $B_t^{(d)} \gets B_t^{(d)} \bigcup tex$
        \State $\widehat{C_{tex}} \gets \widehat{C_{tex}} \setminus tex$
        \State break whle loop
        \EndIf
        \EndIf
        \EndWhile
        
        \EndWhile
        \State $B_t \gets B_t \bigcup B_t^{(d)}$
        \EndFor
        \State \Return $B_t$
        \EndFunction
    \end{algorithmic}
\end{algorithm}

It's worth noting that the texture components and smooth background components in the proposed anomaly detection methods are generated by different basis functions, resulting in differences in data scale. According to statistical results obtained from experiments on realistic datasets in this study, the value range of texture components directly generated by the mathematical formulas in Appendix \ref{appsec:ridgeReg_TBFLP} and \ref{appsec:ridgeReg_ADP} is generally 2 to 3 times higher than that of the smooth background components. However, utilizing the relevant data without processing will cause the loss of texture components during the background subtraction method. The value of the texture component removed from the corresponding region of the overall image data is relatively large, thus polluting the anomaly component obtained from the subtraction with the "texture component removal trace." 

To deal with that problem, the parameter $\phi_{B/T}$ is introduced to balance the directly calculated texture component against the smooth background component. 
\begin{algorithm}[H]
    \caption{Closed Boundary Generation and Interior Area Filling algorithm based on Detected defects}  
    \label{algorithm:ClosedCurveForm}
	\begin{algorithmic}[1] 
		\Require Detected anomalies: $\widehat{C_a}$, Maximum rotation times: $maxRotate$, Maximum distance of pixel neighborhood: $D$, Size of 2D image data: $m\times n$
		\Ensure Boundary knots set: $S_{bound}$, Closed boundary curves set: $S_{curve}$, Closed defects interior pixels set: $S_{area}$
		\Function {ClosedCurveForm}{$\widehat{C_a}, maxRotate, D, m, n$}  
        \State $\widehat{C_a^D} \gets$ \ defect-free \ points \ within \ distance \ $D$ from \ defect \ points \ in \ $\widehat{C_a}$ 
        \State $K \gets$ \ number \ of \ points \ in \ $\widehat{C_a^D}$
		\For {$k \in [1, K]$}
		\State $P_a^D \gets$ \ center \ point \ of $\widehat{C_a^D}$
		\State $\alpha \gets 2\pi / maxRotate$
		\For {$i_R \in [0, maxRotate]$}
		\State $S_{dot}^{i\alpha, k} \gets$ \ all \ defect \ points \ within \ range $(i_R\alpha, (i_R+1)\alpha)$ \ from \ $P_a^D$
		\State $S_{bound}^k \gets \{$ the \ furthest \ point \ $\in S_{dot}^{i\alpha, k}$ \ from $P_a^D\}$
		\EndFor
		\State $S_{curve}^k \gets S_{curve}^k \bigcup f_{B-spline}(S_{bound}^k)$
		\State $i\gets 0, j\gets 0$
		\While {$i<m$}
		\While {$j<n$}
		\State $P_{(i, j)}\gets$ $(i, j)$ \ pixel \ of \ the \ image
		\If {$P_{(i, j)}$ in \ the \ closed \ area \ surrounded \ by $S_{curve}^k$}
		\State $S_{area}^k\gets S_{area}^k \bigcup P_{(i, j)}$
		\EndIf
		\EndWhile
		\EndWhile
		\State $S_{bound}, S_{curve}, S_{area} \gets S_{bound} \bigcup S_{bound}^{k}, S_{curve} \bigcup S_{curve}^{k}, S_{area} \bigcup S_{area}^{k}$
		\EndFor
		\State \Return {$S_{bound}, S_{curve}, S_{area}$}
		\EndFunction   
	\end{algorithmic} 
	\begin{center}
		$f_{B-spline}(X)$ : \ calculate \ closed \ curve \ with \ $X$ \ as \ knots \ based \ on \ B-spline \ function
	\end{center}
\end{algorithm}

%% file: appendix/prove.tex
\section{Mathematical Proof}

\subsection{Proof of Theorem \ref{theorem:composite_quasi_periodicity_constraint} by Cauchy-Schwartz Inequality}
\label{appsec:Cauchyproof}
Theorem \ref{theorem:composite_quasi_periodicity_constraint} can be proved by Inequation \ref{ineq:composite_quasi_periodicity_constraint_prove} with Cauchy-Schwartz Inequality.
    \begin{equation}
        \begin{aligned}
            &{\| \sum_{k=1}^{K}{\beta^{(k)} [{(S^{(k)})}^{\widetilde{t_m}}-{(S_{\widetilde{T^{(k)}}})}^{\widetilde{t_m}} }] }\|^2 
            \le 
            {\| \sum_{k=1}^{K}{\beta^{(k)} [{(S^{(k)})}^{\widetilde{t_m}}-S_{\widetilde{T^{(k)}}} }] }\|^2
            \\ &\le 
            (\sum_{k=1}^{K}{\beta^{(k)}}^2) (\sum_{k=1}^{K}{\| {(S^{(k)})}^{\widetilde{t_m}}-S_{\widetilde{T^{(k)}}} \|^2}) 
            \le 
            (\sum_{k=1}^{K}{\beta^{(k)}}^2) (\sum_{k=1}^{K}{\| {(S^{(k)})}^{\widetilde{t_m}} + S^{\widetilde{t_{m-1}^{(k)}}} -S_{\widetilde{T^{(k)}}} \|^2}) 
            \\ &\le
            (\sum_{k=1}^{K}{\beta^{(k)}}^2) [\sum_{k=1}^{K}(\| {(S^{(k)})}^{\widetilde{t_m}} \|_2 + \| S^{\widetilde{t_{m-1}^{(k)}}} -S_{\widetilde{T^{(k)}}} \|_2)^2] 
            \\ &= 
            (\sum_{k=1}^{K}{\beta^{(k)}}^2) [\sum_{k=1}^{K}(\| {(S^{(k)})}^{\widetilde{t_m}} \|_2 + \sigma^{(k)})^2]
            \le \sigma^2 ,
        \end{aligned}
        \label{ineq:composite_quasi_periodicity_constraint_prove}
    \end{equation}

\subsection{Estimating \texorpdfstring{$\theta, C_{tex}$}{θ} in the Texture Basis Function Learning Process} 
\label{appsec:ridgeReg_TBFLP}

\hspace{1em} The optimization problem is:
\begin{equation}
    \begin{aligned}
    & \underset{\theta}{\operatorname{argmin}} \quad \|e\|^2+\lambda \theta^T R\theta + \gamma \Vert C_{tex}\Vert_1 \\ 
    & subject \: to \quad Y=B\theta+C_{tex}+e \\
    \end{aligned}
\end{equation}

\textbf{Known $Y, B, C_{tex}, R$, given parameter $\lambda$, the estimator of $\theta$ is as follows:}
\begin{equation}
    \begin{aligned}
    & \frac{\partial({\|Y-B\theta-C_{tex}\|}^2+\lambda\theta^T R\theta +\gamma \Vert C_{tex}\Vert_1) }{\partial\theta} = 0 \\
    & \Longleftrightarrow \frac{\partial((Y-B\theta-C_{tex} )^T (Y-B\theta-C_{tex}  )+\lambda\theta^T R\theta)}{\partial\theta}=0 \\
    & \Longleftrightarrow \frac{\partial(\theta^T B^T B\theta-2\theta^T B^T\cdot(Y-C_{tex} ))}{\partial\theta}+\lambda(R+R^T )\theta=0 \\
    & \Longleftrightarrow 2(B^T B\theta-B^T\cdot(Y-C_{tex} )+\lambda R\theta)=0 \\
    & \Longleftrightarrow \hat{\theta}=(B^T B+\lambda R)^{-1} B^T\cdot(Y-C_{tex} )
    \end{aligned}
\end{equation}
\newline

Known $Y, B, \theta, R$, given parameter $\gamma$, the estimator of $C_{tex}$ is as follows:
\begin{equation}
    \begin{aligned}
    & \frac{\partial({\|Y-B\theta-C_{tex}\|}^2+\lambda\theta^T R\theta + \gamma \Vert C_{tex}\Vert_1) }{\partial C_{tex}} = 0 \\
    & \Longleftrightarrow \frac{\partial((Y-B\theta-C_{tex} )^T (Y-B\theta-C_{tex}  )+\gamma \Vert C_{tex}\Vert_1)}{\partial C_{tex}}=0 \\
    & \Longleftrightarrow \frac{\partial(C_{tex}^T C_{tex} - 2C_{tex}^T\cdot(Y-B\theta))}{\partial C_{tex}}+\gamma\cdot sgn(C_{tex})=0 \\
    & \Longleftrightarrow 2(C_{tex}-(Y-B\theta ) ) +\gamma\cdot sgn(C_{tex})=0 \\
    & \Longleftrightarrow \widehat{C_{tex}}=(Y-B\theta) - \frac{\gamma}{2}sgn(C_{tex})
    \end{aligned}
\end{equation}
where the mathematical expression of $sgn(C_{tex})$ is as follows:
\begin{equation}
    \begin{aligned}
        & sgn(C_{tex})=sgn(Y-B\theta)-\frac{\gamma}{2} sgn(C_{tex}) \\
        & \Longleftrightarrow  sgn(C_{tex})=\frac{sgn(Y-B\theta)}{1+\frac{\gamma}{2}} \\
        & \overset{0<\gamma\ll1}{\Longleftrightarrow} 
        sgn(C_{tex})=sgn(Y-B\theta)
    \end{aligned}
\end{equation}

\textbf{Therefore, the estimator of $C_{tex}$ with $\theta$ known in advance is as follows:}
\begin{equation}
    \widehat{C_{tex}}=(Y-B\theta) - \frac{\gamma}{2}sgn(C_{tex})=(Y-B\theta) - \frac{\gamma}{2}sgn(Y-B\theta)
\end{equation}

\subsection{Estimating \texorpdfstring{$\theta, \theta_t, C_a$}{θ, } in the Anomaly Detection Process} 
\label{appsec:ridgeReg_ADP}

\hspace{1em} The optimization problem is:
\begin{equation}
    \begin{aligned}
    & \underset{\theta, \theta_t}{\operatorname{argmin}} \quad \|e\|^2+\lambda \theta^T R\theta +\gamma\|\theta_t\|_1 + \eta \Vert C_a\Vert_1\\ 
    & subject \: to \quad Y=B\theta+B_t\theta_t+C_a+e \\
    \end{aligned}
\end{equation}

\textbf{Known $Y, B, B_t, \theta_t, C_a, R$, given parameter $\lambda$, the estimator of $\theta$ is as follows:}
\begin{equation}
    \begin{aligned}
    & \frac{\partial({\|Y-B\theta-B_t\theta_t-C_a\|}^2+\lambda\theta^T R\theta)+\gamma\|\theta_t\|_1 + \eta \Vert C_a\Vert_1}{\partial\theta} = 0 \\
    & \Longleftrightarrow \frac{\partial((Y-B\theta-B_t\theta_t-C_a )^T (Y-B\theta-B_t\theta_t-C_a  )+\lambda\theta^T R\theta)}{\partial\theta}=0 \\
    & \Longleftrightarrow \frac{\partial(\theta^T B^T B\theta-2\theta^T B^T\cdot(Y-B_t\theta_t-C_a ))}{\partial\theta}+\lambda(R+R^T )\theta=0 \\
    & \Longleftrightarrow 2(B^T B\theta-B^T\cdot(Y-B_t\theta_t-C_a)+\lambda R\theta)=0 \\
    & \Longleftrightarrow \hat{\theta}=(B^T B+\lambda R)^{-1} B^T\cdot(Y-B_t\theta_t-C_a) )
    \end{aligned}
\end{equation}
\newline

Known $Y, B, B_t, C_a, \theta$, under the condition of $B_t{B_t}^T=I, 0<\gamma\ll1$, the solution of estimation of $\theta_t$ is as follows:
\begin{equation}
    \begin{aligned}
    & \frac{\partial({\|Y-B\theta-B_t\theta_t-C_a\|}^2+\lambda\theta^T R\theta)+\gamma\|\theta_t\|_1 + \eta \Vert C_a\Vert_1}{\partial\theta_t} = 0 \\
    & \Longleftrightarrow \frac{\partial((Y-B\theta-B_t\theta_t-C_a )^T (Y-B\theta-B_t\theta_t-C_a  )+\gamma\|\theta_t\|_1)}{\partial\theta_t}=0 \\
    & \Longleftrightarrow \frac{\partial(\theta_t^T B_t^T B_t\theta_t-2\theta_t^T B_t^T\cdot(Y-B\theta-C_a ))}{\partial\theta_t}+\gamma\cdot sgn(\theta_t)=0 \\
    & \Longleftrightarrow 2(B_t^T B_t\theta_t-B_t^T\cdot(Y-B\theta-C_a))+\gamma\cdot sgn(\theta_t)=0 \\
    & \overset{B_t{B_t}^T=I}{\Longleftrightarrow} \hat{\theta_t}=B_t^T\cdot(Y-B\theta-C_a) - \frac{\gamma}{2}sgn(\theta_t)
    \end{aligned}
\end{equation}
where the mathematical expression of $sgn(\theta_t)$ is as follows:
\begin{equation}
    \begin{aligned}
    & sgn(\theta_t)=sgn(B_t^T\cdot(Y-B\theta-C_a))-\frac{\gamma}{2} sgn(\theta_t) \\
    & \Longleftrightarrow  sgn(\theta_t)=\frac{sgn(B_t^T\cdot(Y-B\theta-C_a))}{1+\frac{\gamma}{2}} \\
    & \overset{0<\gamma\ll1}{\Longleftrightarrow} 
    sgn(\theta_t)=sgn(B_t^T\cdot(Y-B\theta-C_a))
    \end{aligned}
\end{equation}

\textbf{Therefore the estimator of $\theta_t$ is:}
\begin{equation}
    \hat{\theta_t}=B_t^T\cdot(Y-B\theta-C_a) - \frac{\gamma}{2}sgn(B_t^T\cdot(Y-B\theta-C_a))
\end{equation}
\newline

Known $Y, B, B_t, \theta, \theta_t$, the estimator of $C_a$ is as follows:
\begin{equation}
    \begin{aligned}
    & \frac{\partial({\|Y-B\theta-B_t\theta_t-C_a\|}^2+\lambda\theta^T R\theta)+\gamma\|\theta_t\|_1 + \eta \Vert C_a\Vert_1}{\partial C_a} = 0 \\
    & \Longleftrightarrow \frac{\partial((Y-B\theta-B_t\theta_t-C_a )^T (Y-B\theta-B_t\theta_t-C_a  )+ \eta \Vert C_a\Vert_1}{\partial C_a}=0 \\
    & \Longleftrightarrow \frac{\partial(C_a^TC_a - 2C_a^T \cdot(Y-B\theta-B_t\theta_t ))}{\partial C_a} + \eta \Vert C_a\Vert_1 =0 \\
    & \Longleftrightarrow 2(C_a-(Y-B\theta-B_t\theta_t)) + \eta \Vert C_a\Vert_1 =0 \\
    & \Longleftrightarrow \widehat{C_a}=(Y-B\theta-B_t\theta_t) - \frac{\eta}{2}sgn(C_a)
    \end{aligned}
\end{equation}
where the mathematical expression of $sgn(C_a)$ is as follows:
\begin{equation}
    \begin{aligned}
    & sgn(C_a)=sgn(Y-B\theta-B_t\theta_t) - \frac{\eta}{2}sgn(C_a) \\
    & \Longleftrightarrow  sgn(C_a)=\frac{sgn(Y-B\theta-B_t\theta_t)}{1+\frac{\eta}{2}} \\
    & \overset{0<\eta\ll1}{\Longleftrightarrow} 
    sgn(C_a)=sgn(Y-B\theta-B_t\theta_t)
    \end{aligned}
\end{equation}

\textbf{Therefore the estimator of $C_a$ is:}
\begin{equation}
    \widehat{C_a}=(Y-B\theta-B_t\theta_t) - \frac{\eta}{2}sgn(C_a) = (Y-B\theta-B_t\theta_t) - \frac{\eta}{2}sgn(Y-B\theta-B_t\theta_t)
\end{equation}

%% file: appendix/figures.tex
\subsection{Supplementary Figures}
\begin{figure}[H]
    \centering
    \includegraphics[width=0.8\textwidth]{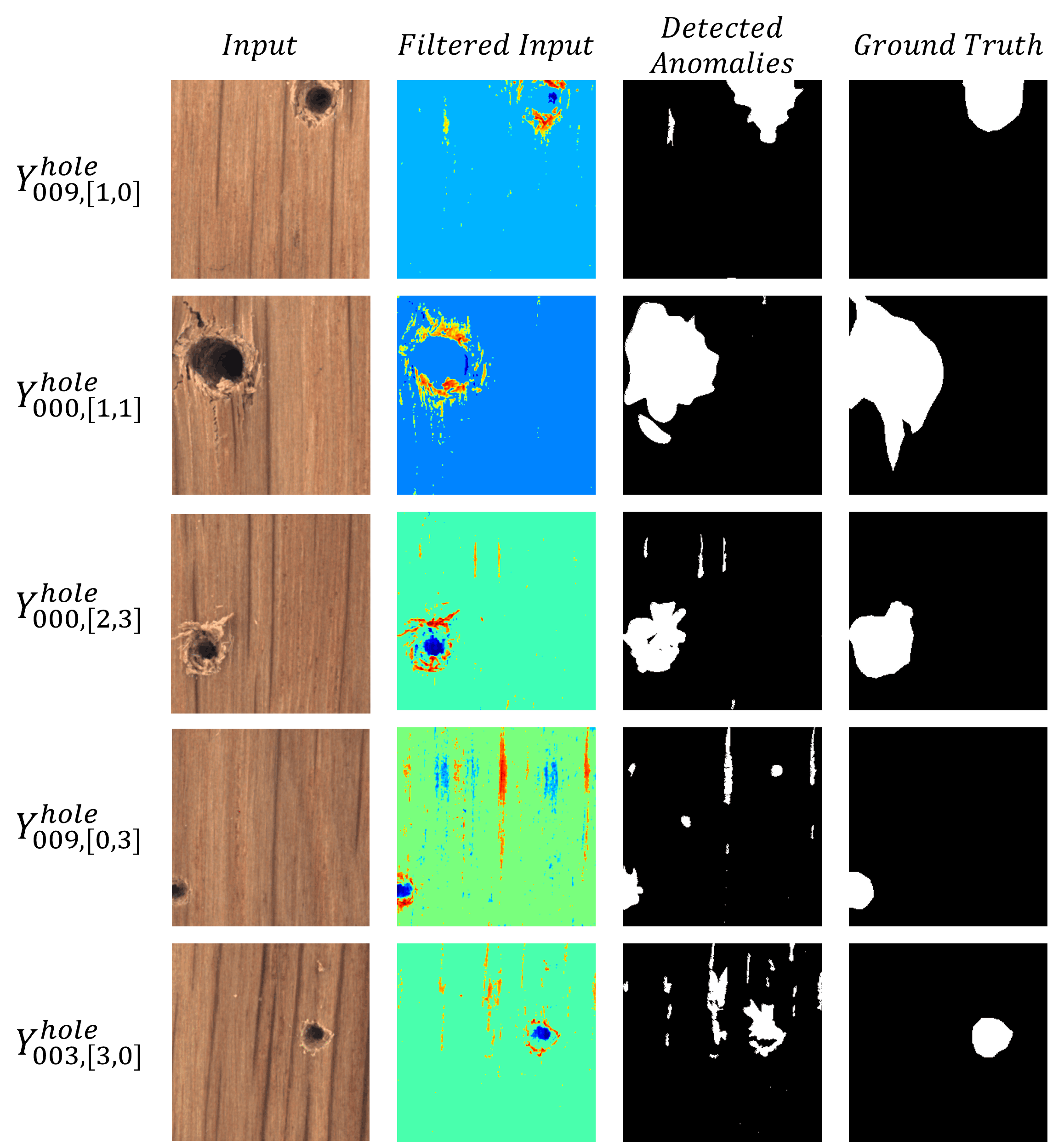}
    \caption{Visualization of five best detection records ranking based on TPR}
    \label{5_FiveBest}
\end{figure}
\begin{figure}[h]
    \centering
    \includegraphics[width=0.8\textwidth]{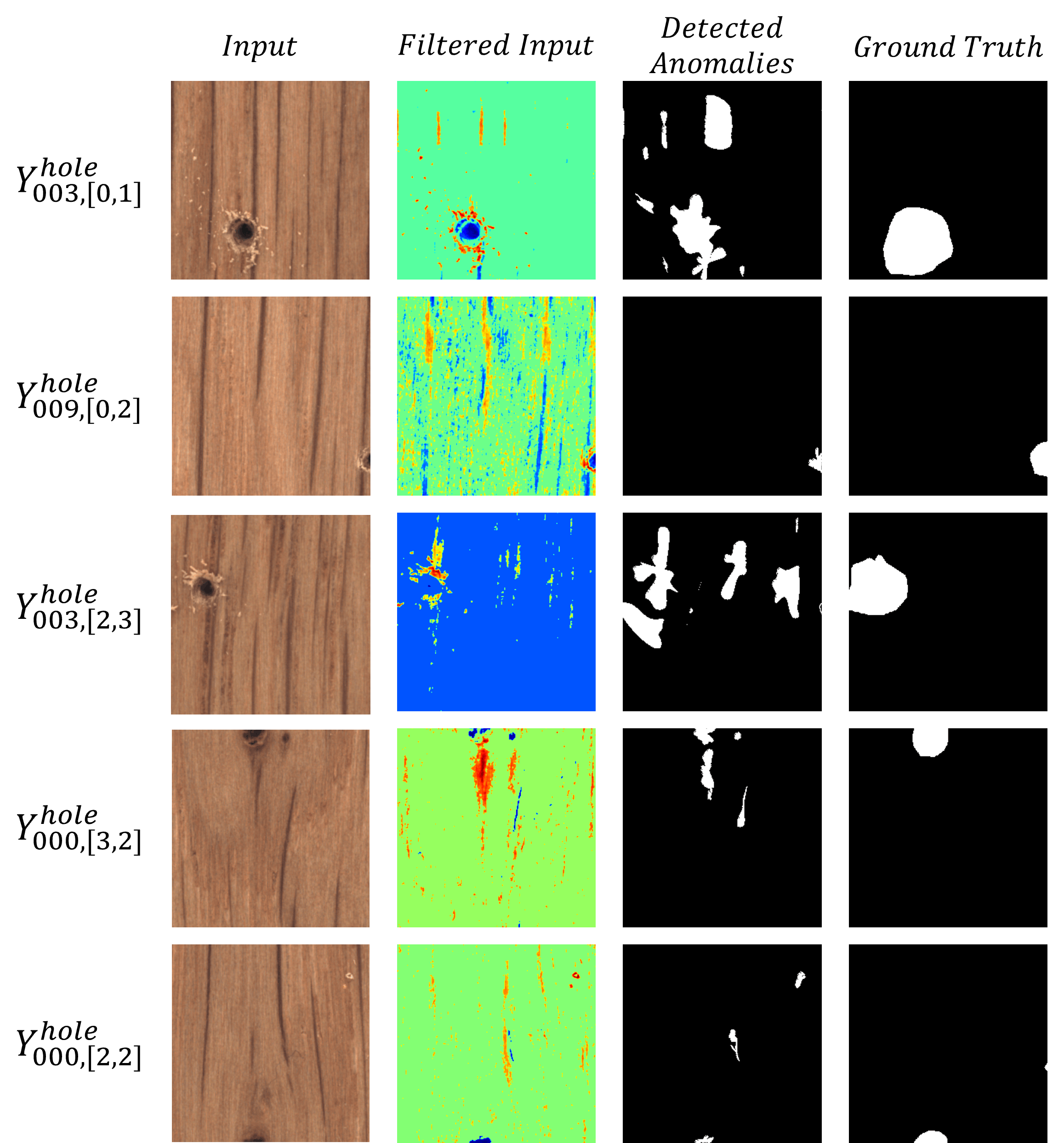}
    \caption{Visualization of five worst detection records ranking based on TPR}
    \label{5_FiveWorst}
\end{figure}

%% file: references.bib
@article{yan2017anomaly,
  author = {Hao Yan and Kamran Paynabar and Jianjun Shi},
  title = {Anomaly Detection in Images with Smooth Background via Smooth-sparse Decomposition},
  journal = {Technometrics},
  volume = {59},
  number = {1},
  pages = {102-114},
  year = {2017},
  doi = {10.1080/00401706.2015.1102764},
}

@article{yan2018real,
  author = {Hao Yan and Kamran Paynabar and Jianjun Shi},
  title = {Real-Time Monitoring of High-Dimensional Functional Data Streams via Spatio-Temporal Smooth Sparse Decomposition},
  journal = {Technometrics},
  volume = {59},
  number = {1},
  pages = {102-114},
  year = {2018},
  doi = {10.1080/00401706.2017.1346522},
}

@article{candes2011robust,
  author = {Emmanuel J. Candès and Li, Xiaodong and Yi Ma and John Wright},
  title = {Robust Principal Component Analysis?},
  journal = {Journal of the ACM (JACM)},
  volume = {58},
  number = {3},
  pages = {1-37},
  year = {2011},
  doi = {10.48550/arXiv.0912.3599},
}

@article{tsai2003automated,
  author = {Du-Ming Tsai and Tse-Yun Huang},
  title = {Automated Surface Inspection for Statistical Textures},
  journal = {Image and Vision Computing},
  volume = {21},
  number = {4},
  pages = {307-323},
  year = {2003},
  doi = {10.1016/S0262-8856(03)00007-6},
}

@article{wilts2000auto,
  author = {Wiltschi, Klaus and Pinz, Axel and Lindeberg, Tony},
	journal = {Machine Vision and Applications},
	month = {10},
	number = {3},
	pages = {113--128},
	title = {{An automatic assessment scheme for steel quality inspection}},
	volume = {12},
	year = {2000},
	doi = {10.1007/s001380050130},
	url = {https://doi.org/10.1007/s001380050130},
}

@inproceedings{mehmood2010anomaly,
  author = {Mehmood, Asif and Nasrabadi, Nasser M.},
  title = {Anomaly Detection in Wavelet Domain for Long-wave FLIR Imagery},
  booktitle = {Automatic Target Recognition XX Acquisition, Tracking, Pointing, and Laser Systems Technologies XXIV and Optical Pattern Recognition XXI},
  volume = {7696},
  pages = {76960S},
  year = {2010},
  doi = {10.1117/12.850211},
}

@article{emmanuel1999ridge,
  author = {Candès, Emmanuel J. and Donoho, David L.},
	journal = {Philosophical Transactions of the Royal Society A Mathematical Physical and Engineering Sciences},
	month = {9},
	number = {1760},
	pages = {2495--2509},
	title = {{Ridgelets: a key to higher-dimensional intermittency?}},
	volume = {357},
	year = {1999},
	doi = {10.1098/rsta.1999.0444},
	url = {https://doi.org/10.1098/rsta.1999.0444},
}

@article{starck2001curvelet,
  author={Starck, None Jean-Luc and Candes, E.J. and Donoho, D.L.},
  journal={IEEE Transactions on Image Processing}, 
  title={The curvelet transform for image denoising}, 
  year={2002},
  volume={11},
  number={6},
  pages={670-684},
  doi={10.1109/TIP.2002.1014998}
}

@article{do2005contourlet,
  author = {Do, M. N. and Vetterli, M.},
  title = {The Contourlet Transform: An Efficient Directional Multiresolution Image Representation},
  journal = {IEEE Transactions on Image Processing},
  volume = {14},
  number = {12},
  pages = {2091-2106},
  year = {2005},
  doi = {10.1109/tip.2005.859376},
}

@article{hma2006HoughTrans,
  author = {Van Der Werff, H.M.A. and Bakker, W.H. and Van Der Meer, F.D. and Siderius, W.},
  title = {Combining Spectral Signals and Spatial Patterns Using Multiple Hough Transforms: An Application for Detection of Natural Gas Seepages},
  journal = {Computers \& Geosciences},
  volume = {32},
  number = {9},
  pages = {1334-1343},
  year = {2006},
  doi = {10.1016/j.cageo.2005.12.003},
}

@inproceedings{cao2022semiknowledge,
  author = {Cao, Yunkang and Song, Yanan and Xu, Xiaohao and Li, Shuya and Yu, Yuhao and Zhang, Yifeng and Shen, Weiming},
  title = {Semi-supervised Knowledge Distillation for Tiny Defect Detection},
  booktitle = {IEEE 25th International Conference on Computer Supported Cooperative Work in Design (CSCWD)},
  pages = {1010-1015},
  year = {2022},
  doi = {10.1109/CSCWD54268.2022.9776026},
}

@article{bergmann2021mvtec,
  author = {Bergmann, Paul and Batzner, Kilian and Fauser, Michael and Sattlegger, David and Steger, Carsten},
	journal = {International Journal of Computer Vision},
	month = {1},
	number = {4},
	pages = {1038--1059},
	title = {{The MVTEC Anomaly Detection Dataset: A comprehensive Real-World dataset for unsupervised anomaly detection}},
	volume = {129},
	year = {2021},
	doi = {10.1007/s11263-020-01400-4},
	url = {https://doi.org/10.1007/s11263-020-01400-4},
}

@inproceedings{tayeh2020distanceAD,
  author = {Tayeh, Tareq and Aburakhia, Sulaiman and Myers, Ryan and Shami, Abdallah},
  title = {Distance-Based Anomaly Detection for Industrial Surfaces Using Triplet Networks},
  booktitle = {2020 11th IEEE Annual Information Technology, Electronics and Mobile Communication Conference (IEMCON)},
  pages = {0372-7},
  year = {2020},
  doi = {10.1109/IEMCON51383.2020.9284921},
}

@inproceedings{li2021ADviaSOM,
  author = {Ning Li and Kaitao Jiang and Zhiheng Ma and Xing Wei and Xiaopeng Hong and Yihong Gong},
  title = {Anomaly Detection via Self-organizing Map},
  booktitle = {2021 IEEE International Conference on Image Processing (ICIP)},
  pages = {974-978},
  year = {2021},
  doi = {10.1109/ICIP42928.2021.9506433},
}

@article{desforges1998Parzen,
author = {Desforges, M J and Jacob, P J and Cooper, J E},
title ={Applications of probability density estimation to the detection of abnormal conditions in engineering},
journal = {Proceedings of the Institution of Mechanical Engineers, Part C: Journal of Mechanical Engineering Science},
volume = {212},
number = {8},
pages = {687-703},
year = {1998},
doi = {10.1243/0954406981521448},
URL = {https://doi.org/10.1243/0954406981521448},
eprint = {https://doi.org/10.1243/0954406981521448}
}

@article{Veracini2011Nonparametric,
  author={Veracini, Tiziana and Matteoli, Stefania and Diani, Marco and Corsini, Giovanni},
  journal={IEEE Geoscience and Remote Sensing Letters}, 
  title={Nonparametric Framework for Detecting Spectral Anomalies in Hyperspectral Images}, 
  year={2011},
  volume={8},
  number={4},
  pages={666-670},
  doi={10.1109/LGRS.2010.2099103}}

@inproceedings{wyatt2022anoDDPM,
  author={Wyatt, Julian and Leach, Adam and Schmon, Sebastian M. and Willcocks, Chris G.},
  booktitle={2022 IEEE/CVF Conference on Computer Vision and Pattern Recognition Workshops (CVPRW)}, 
  title={AnoDDPM: Anomaly Detection with Denoising Diffusion Probabilistic Models using Simplex Noise}, 
  year={2022},
  volume={},
  number={},
  pages={649-655},
  doi={10.1109/CVPRW56347.2022.00080}}

@ARTICLE{Hu2020localKDE,
  author={Hu, Weiming and Gao, Jun and Li, Bing and Wu, Ou and Du, Junping and Maybank, Stephen},
  journal={IEEE Transactions on Knowledge and Data Engineering}, 
  title={Anomaly Detection Using Local Kernel Density Estimation and Context-Based Regression}, 
  year={2020},
  volume={32},
  number={2},
  pages={218-233},
  doi={10.1109/TKDE.2018.2882404}}

@ARTICLE{Tu2020HADusingDWD,
  author={Tu, Bing and Yang, Xianchang and Zhou, Chengle and He, Danbing and Plaza, Antonio},
  journal={IEEE Transactions on Geoscience and Remote Sensing}, 
  title={Hyperspectral Anomaly Detection Using Dual Window Density}, 
  year={2020},
  volume={58},
  number={12},
  pages={8503-8517},
  doi={10.1109/TGRS.2020.2988385}}

@ARTICLE{Kareth2022AD_HMM,
  author={León-López, Kareth M. and Mouret, Florian and Arguello, Henry and Tourneret, Jean-Yves},
  journal={IEEE Transactions on Geoscience and Remote Sensing}, 
  title={Anomaly Detection and Classification in Multispectral Time Series Based on Hidden Markov Models}, 
  year={2022},
  volume={60},
  number={},
  pages={1-11},
  doi={10.1109/TGRS.2021.3101127}}

@ARTICLE{Zavrtanik2021reconstruction,
    author = {Vitjan Zavrtanik and Matej Kristan and Danijel Skočaj},
    title = {Reconstruction by inpainting for visual anomaly detection},
    journal = {Pattern Recognition},
    volume = {112},
    pages = {107706},
    year = {2021},
    issn = {0031-3203},
    doi = {https://doi.org/10.1016/j.patcog.2020.107706},
    url = {https://www.sciencedirect.com/science/article/pii/S0031320320305094}
}

@ARTICLE{zhou2019anomalynet,
  author={Zhou, Joey Tianyi and Du, Jiawei and Zhu, Hongyuan and Peng, Xi and Liu, Yong and Goh, Rick Siow Mong},
  journal={IEEE Transactions on Information Forensics and Security}, 
  title={AnomalyNet: An Anomaly Detection Network for Video Surveillance}, 
  year={2019},
  volume={14},
  number={10},
  pages={2537-2550},
  doi={10.1109/TIFS.2019.2900907}}

@article{li_anomalydetectionbased_2018,
  title = {Anomaly Detection Based on Sparse Coding with Two Kinds of Dictionaries},
  author = {Shifeng Li and Chunxiao Liu and Yuqiang Yang},
  year = {2018},
  month = jul,
  journal = {Signal, Image and Video Processing},
  volume = {12},
  number = {5},
  pages = {983--989},
  issn = {1863-1703, 1863-1711},
  doi = {10.1007/s11760-018-1243-7},
  urldate = {2025-01-13},
  langid = {english}
}

@article{Lahoti02102021,
author = {Geet Lahoti and Jialei Chen and Xiaowei Yue and Hao Yan and Chitta Ranjan and Zhen Qian and Chuck Zhang and Ben Wang},
title = {Image decomposition-based sparse extreme pixel-level feature detection model with application to medical images},
journal = {IISE Transactions on Healthcare Systems Engineering},
volume = {11},
number = {4},
pages = {338--354},
year = {2021},
publisher = {Taylor \& Francis},
doi = {10.1080/24725579.2021.1910599},
URL = { 
    
        https://doi.org/10.1080/24725579.2021.1910599    

},
eprint = {        https://doi.org/10.1080/24725579.2021.1910599

}

}

@ARTICLE{Albahar1,
  author={AlBahar, Areej and Kim, Inyoung and Yue, Xiaowei},
  journal={IEEE Transactions on Automation Science and Engineering}, 
  title={A Robust Asymmetric Kernel Function for Bayesian Optimization, With Application to Image Defect Detection in Manufacturing Systems}, 
  year={2022},
  volume={19},
  number={4},
  pages={3222-3233},
  doi={10.1109/TASE.2021.3114157}}

@article{Gao02092023,
author = {Yuanyuan Gao and Xinming Wang and Junbo Son and Xiaowei Yue and Jianguo Wu},
title = {Hierarchical modeling of microstructural images for porosity prediction in metal additive manufacturing via two-point correlation function},
journal = {IISE Transactions},
volume = {55},
number = {9},
pages = {957--969},
year = {2023},
publisher = {Taylor \& Francis},
doi = {10.1080/24725854.2022.2115593},


URL = { 
    
        https://doi.org/10.1080/24725854.2022.2115593
    
    

},
eprint = { 
    
        https://doi.org/10.1080/24725854.2022.2115593
    
    

}

}

@article{Wang04052022,
author = {Yinan Wang and Weihong "Grace" Guo and Xiaowei Yue},
title = {Tensor decomposition to compress convolutional layers in deep learning},
journal = {IISE Transactions},
volume = {54},
number = {5},
pages = {481--495},
year = {2022},
publisher = {Taylor \& Francis},
doi = {10.1080/24725854.2021.1894514},


URL = { 
    
        https://doi.org/10.1080/24725854.2021.1894514
    
    

},
eprint = { 
    
        https://doi.org/10.1080/24725854.2021.1894514
    
    

}

}

@INPROCEEDINGS{MSEC2019,
    author = {Yue, Xiaowei},
    title = {Data Decomposition for Analytics of Engineering Systems: Literature Review, Methodology Formulation, and Future Trends.
             {V}olume 1: Additive Manufacturing; Manufacturing Equipment and Systems; Bio and Sustainable Manufacturing},
    series = {International Manufacturing Science and Engineering Conference},
    pages = {V001T02A011},
    year = {2019},
    month = {06},
    doi = {10.1115/MSEC2019-2945},
    url = {https://doi.org/10.1115/MSEC2019-2945},
}

@article{xu2025change,
  author = {Xu, Ruiyu and Song, Zheren and Wu, Jianguo and Wang, Chao and Zhou, Shiyu},
  title = {Change-Point Detection with Deep Learning: A Review},
  journal = {Frontiers of Engineering Management},
  year = {2025},
  volume = {12},
  number = {1},
  pages = {154--176}
}
